\pdfoutput=1
\documentclass[11pt,usenames,dvipsnames]{article}

\usepackage{emnlp2022}

\usepackage{subfigure}

\usepackage{emnlp2022}
\usepackage{times}
\usepackage{latexsym}
\usepackage{tabularx}
\usepackage{multirow}
\usepackage{booktabs} % To thicken table lines
\usepackage{enumitem}
\usepackage{amsmath}
\usepackage{inconsolata}
\usepackage{graphicx}

\usepackage[utf8]{inputenc} % allow utf-8 input
\usepackage[T1]{fontenc}    % use 8-bit T1 fonts
\usepackage{hyperref}       % hyperlinks
\usepackage{url}            % simple URL typesetting
\usepackage{booktabs}       % professional-quality tables
\usepackage{amsfonts}       % blackboard math symbols
\usepackage{nicefrac}       % compact symbols for 1/2, etc.
\usepackage{microtype}      % microtypography
\usepackage{xcolor}         % colors
\usepackage{multicol}
\usepackage{transparent}
\usepackage{array}
\usepackage{bm}
\usepackage{smile}

\usepackage{perpage} %the perpage package
\MakePerPage{footnote} %the perpage package command
\usepackage{tabularx}
\usepackage{floatrow, makecell}
\usepackage{wrapfig,lipsum,booktabs}
\usepackage{pifont}% 
\newcommand{\cmark}{\ding{51}}% 
\newcommand{\xmark}{\ding{55}}% 
\definecolor{midnightgreen}{rgb}{0.0, 0.29, 0.33}
\definecolor{chocolate}{rgb}{0.5, 0.2, 0.1}

\definecolor{darkpink}{rgb}{0.91, 0.33, 0.5}

\newcommand{\blue}[1]{\textcolor{blue}{#1}}

\newcommand{\cocobase}{COCO-DR$_{\text{Base}}$}
\newcommand{\cocolarge}{COCO-DR$_{\text{Large}}$}

\newcommand{\model}{COCO-DR}
\newcommand{\marco}{MS MARCO}

\newcommand{\cocondenser}{coCondenser}
\newcommand{\condenser}{Condenser}

\title{COCO-DR: Combating Distribution Shifts  in Zero-Shot Dense Retrieval with Contrastive and Distributionally Robust Learning}

 \author{
   Yue Yu$^{1}$\thanks{\hspace{5pt}Work partly done during Yue's internship at Microsoft.} \quad
 	Chenyan Xiong$^2$ \quad Si Sun$^3$ \quad Chao Zhang$^1$ \quad Arnold Overwijk$^2$ \\
 	 \textsuperscript{1} Georgia Institute of Technology \quad  \textsuperscript{2} Microsoft \quad \textsuperscript{3} Tsinghua University \\ 
 	 \texttt{\{yueyu, chaozhang\}@gatech.edu}, \quad \texttt{s-sun17@mails.tsinghua.edu.cn} \\
 	\texttt{\{chenyan.xiong, arnold.overwijk\}@microsoft.com}
 }    
     
\begin{document}

\maketitle

\begin{abstract}
We present a new zero-shot dense retrieval (ZeroDR) method, COCO-DR, to improve the generalization ability of dense retrieval by combating the distribution shifts between source training tasks and target scenarios.
To mitigate the impact of  document differences,
COCO-DR continues pretraining the language model on the target corpora to adapt to target distributions via COtinuous COtrastive learning.
To prepare for unseen target queries, COCO-DR leverages implicit  Distributionally Robust Optimization (iDRO) to reweight samples from different source query clusters for improving model robustness over rare queries during fine-tuning.
COCO-DR achieves superior average performance on BEIR, the zero-shot retrieval benchmark. 
At BERT$_\text{Base}$ scale, {\cocobase} outperforms other ZeroDR models with 60$\times$ larger size. At BERT$_\text{Large}$ scale, {\cocolarge}  outperforms the giant GPT-3 embedding model which has 500$\times$ more parameters.
Our analysis show the correlation of COCO-DR's effectiveness in combating distribution shifts and improving zero-shot accuracy.
Our code and model can be found at \url{https://github.com/OpenMatch/COCO-DR}.
\end{abstract}

\section{Introduction}

Learning to represent and match queries and documents by embeddings,
dense retrieval (DR) achieves strong performances in scenarios with sufficient training signals~\citep{msmarco, nq}.
However, in many real world scenarios, obtaining relevance labels can be challenging due to the reliance on domain expertise, or even infeasible because of the strict privacy constraints. 
Deploying dense retrieval in these scenarios becomes zero-shot (ZeroDR,~\citet{beir}),
which requires first training DR models on source tasks and then generalizing to target tasks with zero in-domain supervision~\citep{contriever, gtr, cpt}.

ZeroDR poses great challenges to the generalization ability of DR models under the distribution shift between source and target data~\citep{gulrajani2020search,wiles2021fine}, as 
it requires the alignment between queries and their relevant documents in the embedding space.  
% Without guaranteed exact word overlap,
% Matching texts solely via learned embeddings are more prone to distribution shifts by nature.
% Dense retrieval also requires the alignment between queries and their relevant documents in the embedding space. 
It is much harder to generalize than standard classification or ranking tasks, where a robust decision boundary is sufficient~\citep{modir}.
\begin{figure}[t]
\centering
\includegraphics[width=0.9\linewidth]{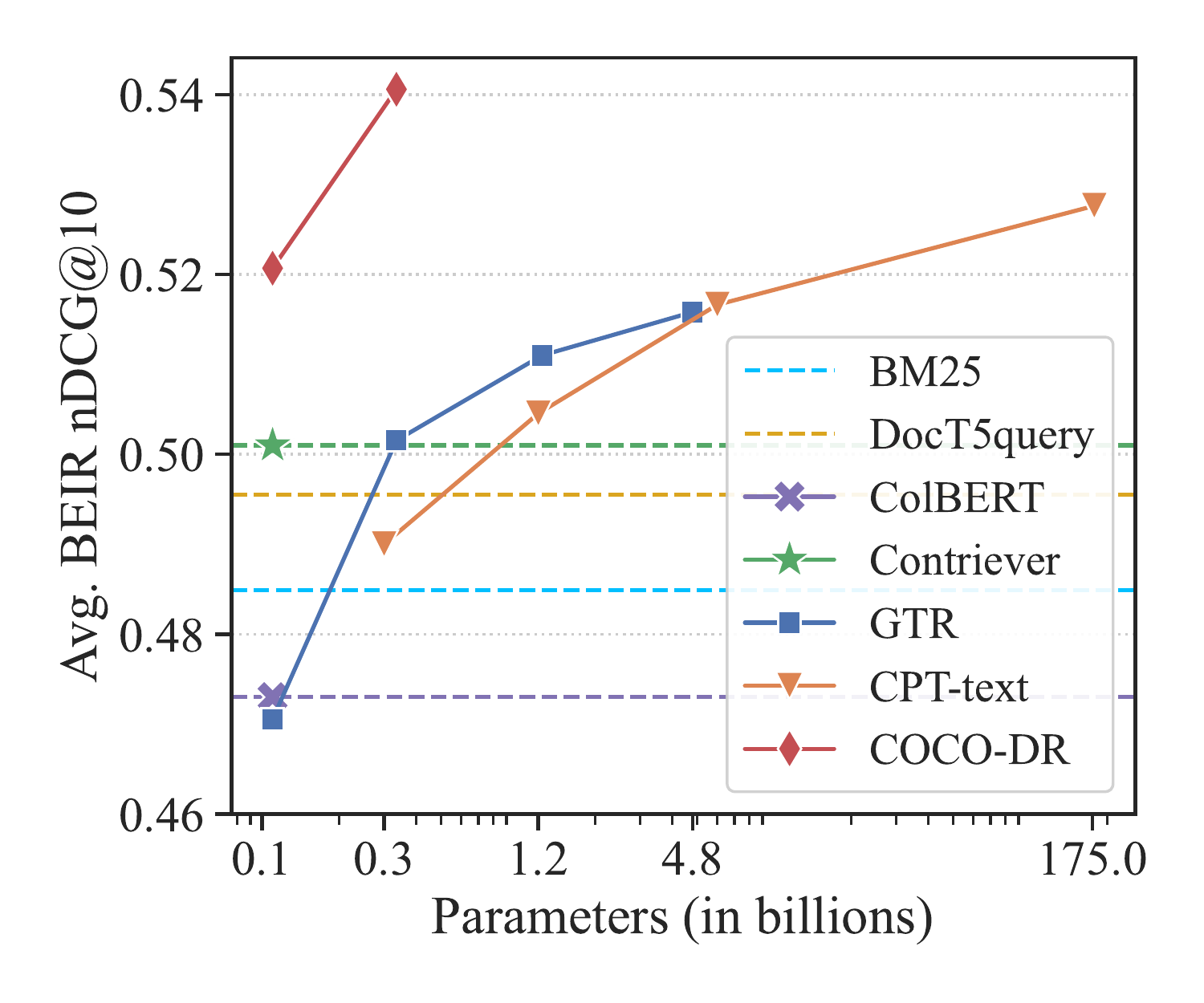}
\vspace{-2ex}
\caption{The average nDCG@10 of {\model} versus large scale models on the 11 BEIR tasks selected in \citet{cpt}.
% {\model} achieves competitive results but with <0.1\% parameters.
X-axis is in log scale.
}
\label{fig:intro-example}
\end{figure}

In this work, we first analyze the distribution shifts in zero-shot dense retrieval.
We illustrate the significant distribution shifts in both query intent and document language from the source to target tasks. After that, we show the strong correlation between the distribution shifts and the reduced  zero-shot accuracy of dense retrieval models, which confirms the 
negative impact of distribution shifts on the generalization ability of dense retrieval. 

We then present \model{}, a ZeroDR model  that combats the distribution shifts between source and target tasks.
In many ZeroDR scenarios, even though relevancy labels or queries are unavailable, the target corpus is often available pre-deploy (otherwise there is nothing to index)~\citep{modir, wang2021gpl}.
We thus design \model{} to perform COntinuous COntrastive pretraining (COCO) on the target corpora, which treats two text sequences from the same document as positive pairs and sequences from different documents as negative pairs. 
% to push the embeddings of  two text sequences from the same document close while pulling the sequences from different documents apart. 
This enables \model{} to mitigate document distribution shifts by improving the alignment and uniformity of sequence representations for target tasks.
% making the pretraining model learn better capture sequence representations  via contrastive learning.

The distribution shift on the query intent, however, is more challenging as there only exists a few, if any, example queries available under ZeroDR scenarios. 
\model{} introduces an implicit distributionally robust optimization (iDRO) method when fine-tuning on the source retrieval labels.
Specifically, it first clusters the source queries into groups based on their learned embeddings.
Then, it dynamically reweights the losses on these query clusters by using the gradient similarity among groups.
% groups.
%  leverage gradient similarity to measure the similarity between groups dynamically reweigh the group-wise losses to maximize loss gain
% balances the gradient operations  \zc{balances the gradient operations is also unclear, better to slightly expand} 
This improves model robustness on less represented query groups in the source, thus implicitly boosts the generalization ability of the DR model on unseen target queries.

\model{} is conceptually simple but empirically powerful.
On 18 retrieval tasks included in BEIR, the standard ZeroDR benchmark~\citep{beir}, \model{} outperforms state-of-the-art domain adaptation methods~\citep{wang2021gpl} which leverage per-task generated pseudo labels and cross-encoder teachers.
\model{} also outperforms large scale models with orders of magnitude more parameters. As shown in Figure~\ref{fig:intro-example}, at only BERT$_\text{base}$ scale with 110M parameters, \model{} outperforms GTR$_\text{XXL}$~\citep{gtr} and CPT$_\text{L}$~\citep{cpt}, which use  $\sim$50$\times$ more parameters.
At BERT$_\text{Large}$ scale, \model{} surpasses CPT$_\text{XL}$~\citep{cpt}, the largest DR model to date (175B parameters) on its selected tasks, only using 0.17\% of its parameters.
% .

Our analysis confirms that the better generalization ability of \model{} comes from its ability to combat the distribution shifts.
Continuous contrastive learning helps the pretrained model better capture target corpora' sequence representation, leading to better generalization ability of models after fine-tuning. 
Training with iDRO helps \model{} achieve robust performances on source query clusters that share similar search intents to target queries, which then lead to better jgeneralization to corresponding target tasks. 

In the rest of this paper, we discuss related work in Section \ref{sec:related}, analyze the distribution shift in Section \ref{sec:distribution}, and present COCO-DR in Section \ref{sec:method}. Our experiments are discussed in Section \ref{sec:exp} and we conclude in Section \ref{sec:conclusion}.

\section{Related Work}
\label{sec:related}

% To represent query and document in a learned space for better first-stage retrieval is not a recent idea. 
Earlier research has explored various ways to learn representations for retrieval~\citep{deerwester1990indexing, huang2013learning}.
Recently, with pretrained language models~\citep{ict}, hard training negative selection~\citep{dpr, ance}, and retrieval-oriented  pretraining~\citep{seed, cocondenser}, dense retrieval has shown strong advantages over sparse retrieval methods,  although the advantages are more observed in supervised settings than zero-shot scenarios~\citep{beir}.

One research direction to improve zero-shot dense retrieval is bringing in domain adaption techniques.
\citet{modir} employ domain invariant learning to narrow the representation gap between source and target domains. 
\citet{genq} and \citet{wang2021gpl} generate pseudo labels for each target task to train in-domain DR models.
These techniques employ one specially trained retrieval model for each target task and improve zero-shot retrieval accuracy.

Another way to improve ZeroDR is to scale up model size and source training data.
\citet{gtr} and \citet{cpt} leverage models with billions of parameters  (T5-XXL and GPT-3) and large-scale training data to increase the generalization capacity of DR model. 
\citet{contriever} and \citet{xu-etal-2022-laprador} enlarge the size of training data with retrieval-oriented pretraining tasks.
As illustrated in Figure~\ref{fig:intro-example}, 
the benefit of scale follows the scaling law of language models~\citep{kaplan2020scaling}: A linear increment of zero-shot accuracy requires exponentially more training data and model parameters. 

Combining dense models with sparse retrieval yields better zero-shot retrieval performances on BEIR~\cite{splade, xu-etal-2022-laprador}.
The reranking models, using stronger cross-encoders, can be used as teachers to improve the robustness of dense retrieval models~\citep{wang2021gpl}.

More generally speaking, continuous pretraining and distributionally robust optimization (DRO) are two techniques for improving model generalization on other applications.
Continuous pretraining BERT's masked language modeling tasks on target domain corpora have shown benefits on both language tasks~\citep{gururangan2020dont}  and the reranking step of search systems~\citep{wang2021domain}.
The benefits of DRO are more ambivalent~\citep{gulrajani2020search, wiles2021fine} and are more observed when explicit group partitions are available~\cite{oren-etal-2019-distributionally,Sagawa2020Distributionally, zhou2021examining}.

\begin{figure*}[t]
% \vspace{-5pt}
\subfigure[Q, ANCE (BERT)]{
    \centering
     \includegraphics[width=0.23\textwidth]{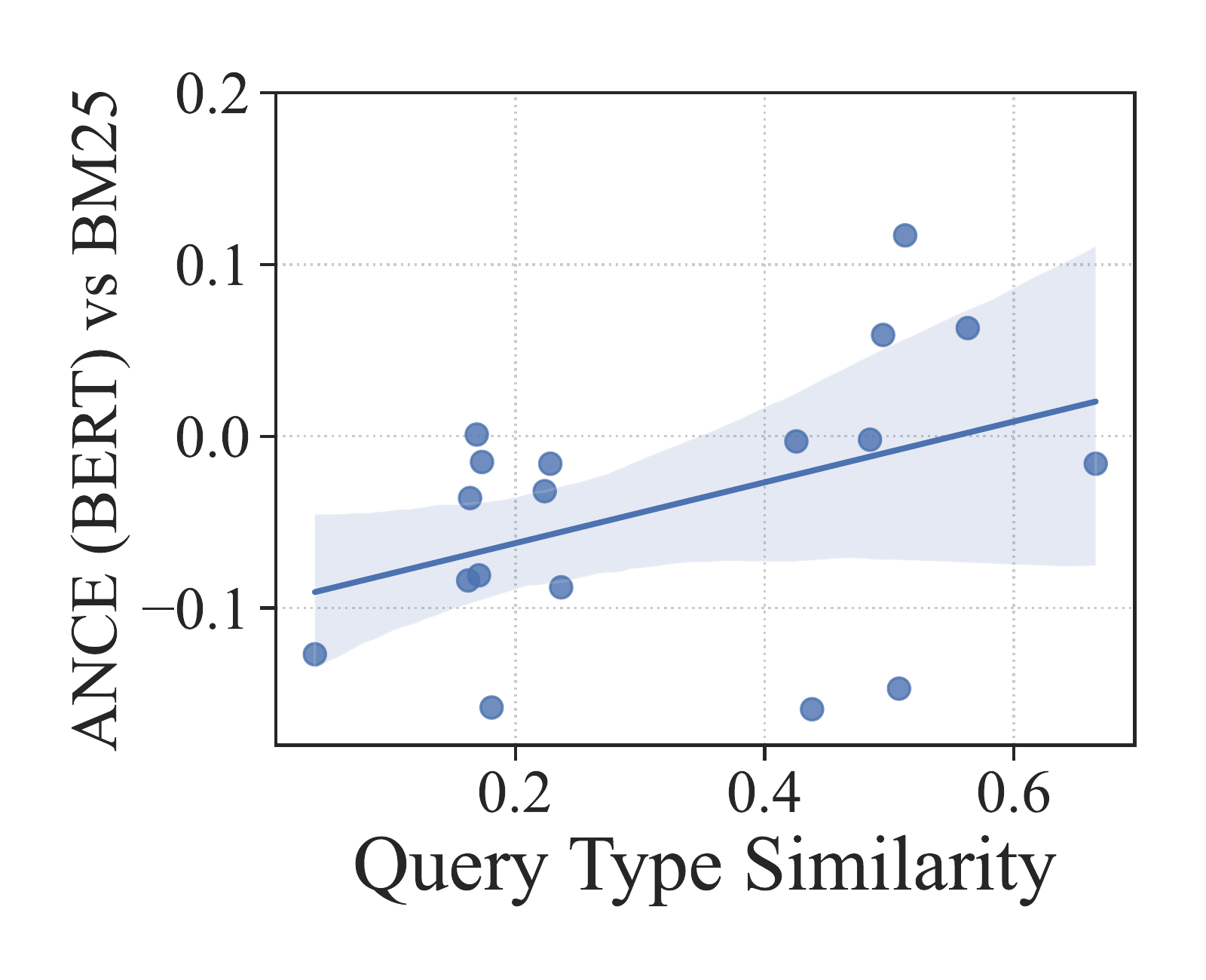}
    \label{fig:ance_q}
    }~
\subfigure[Q, ANCE ({\cocondenser})]{
    \centering
     \includegraphics[width=0.23\textwidth]{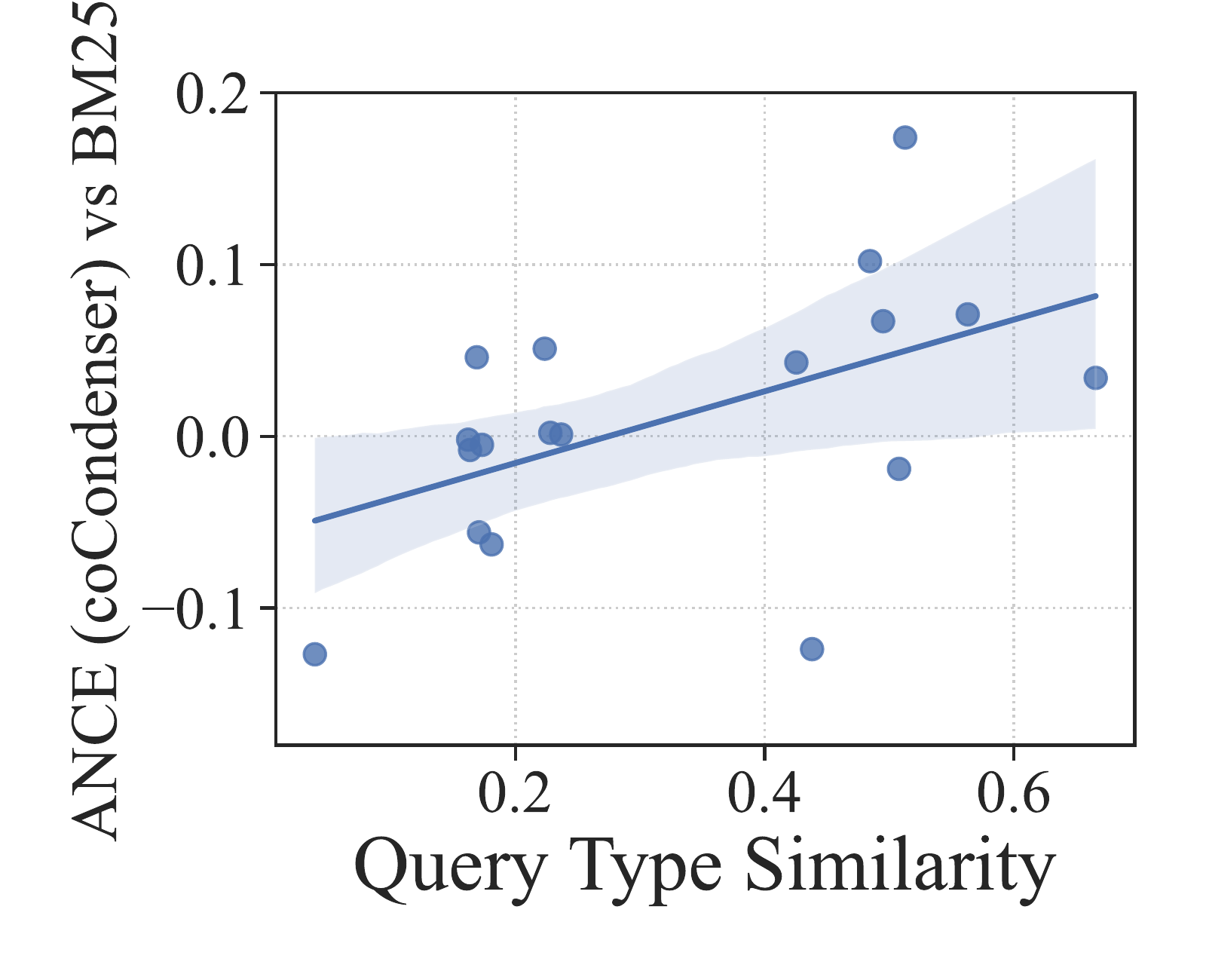}
    \label{fig:cocondenser_q}
    }~
\subfigure[Doc, ANCE (BERT)]{
    \centering
     \includegraphics[width=0.23\textwidth]{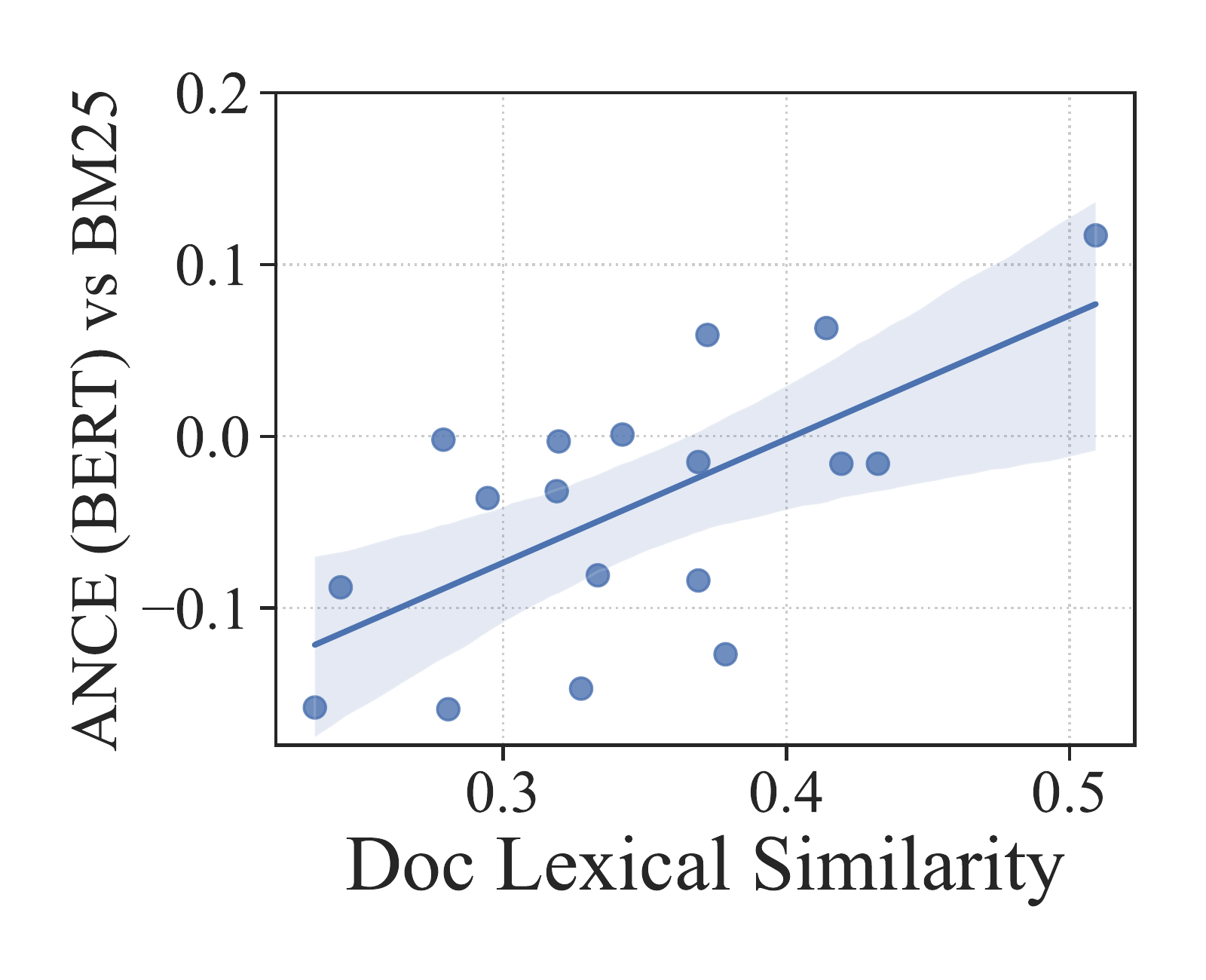}
    \label{fig:ance_d}
    }~ 
\subfigure[Doc, ANCE ({\cocondenser})]{
    \centering
     \includegraphics[width=0.23\textwidth]{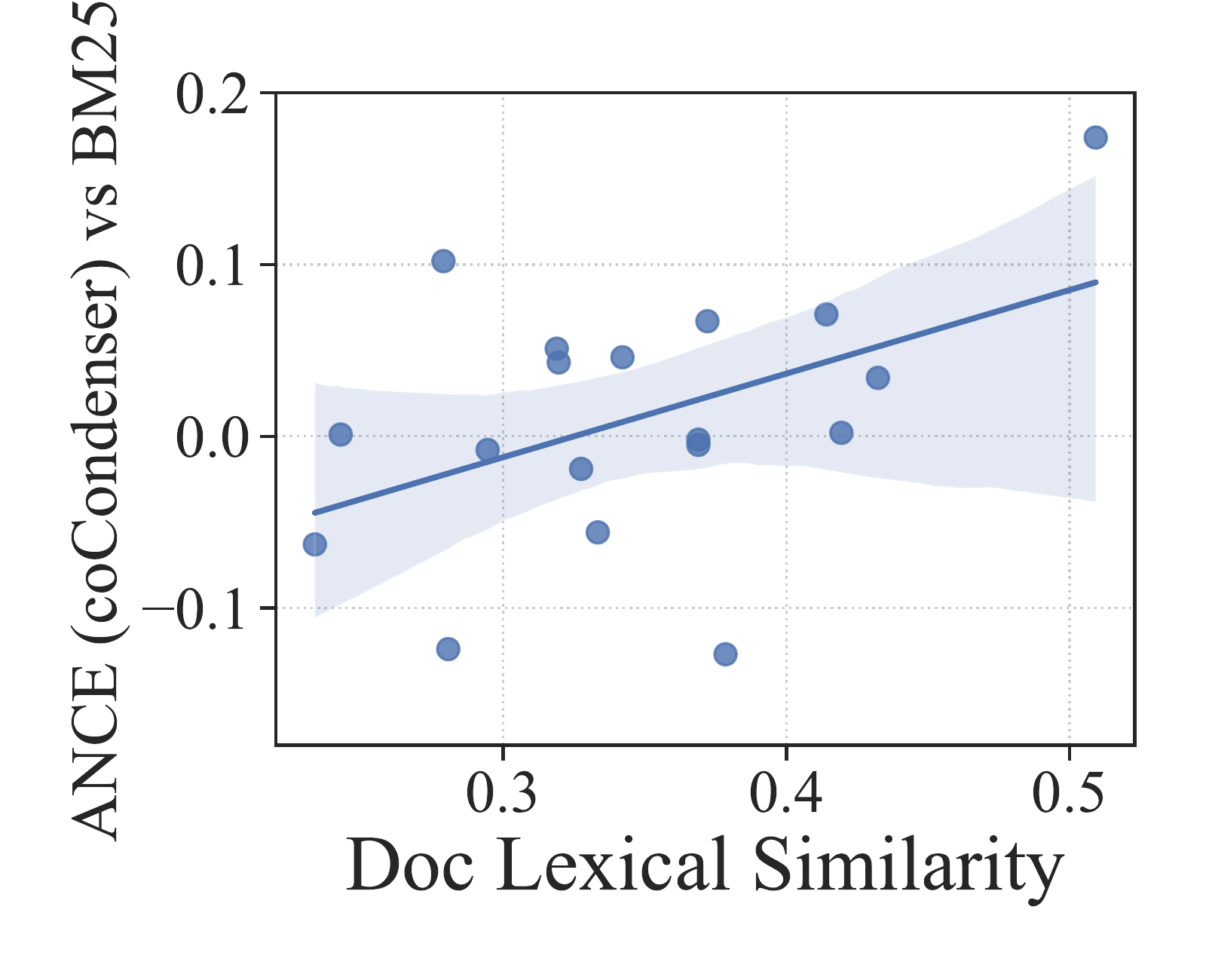}
    \label{fig:cocondenser_d}
    }
    \vspace{-0.5ex}
\caption{Distribution shifts and zero-shot retrieval performances of ANCE trained on {\marco}. X-axes are the similarity between {\marco} and BEIR. Y-axes are NDCG@10 differences on BEIR.
}
\vspace{-0.5ex}
\label{fig:motivation-qdsim}
\end{figure*}

\section{Distribution Shifts in Dense Retrieval}
\label{sec:distribution}

% In this section, we first introduce preliminary information for dense retrieval as well as the ZeroDR setup. Then, we quantify the distribution shift between the source and target tasks for both queries and documents and establish the correlation between distribution shifts and the DR model's zero-shot performance.

In this section, we first introduce the preliminaries of dense retrieval.
Then we discuss the standard zero-shot dense retrieval settings and study the impact of distribution shifts on ZeroDR accuracy.

\subsection{Preliminaries on Dense Retrieval}
In dense retrieval, the query $q$ and document $d$ are represented by \textit{dense} vectors~\citep{huang2013learning} and the relevance score $f(q, d; \theta)$ is often calculated by simple similarity metrics, \emph{e.g.}, dot product~\citep{ict}:
\begin{equation}
f(q, d; \theta) = \left\langle g(q;\theta), g(d;\theta)\right\rangle . \label{eq:dualencoder}
\end{equation}
Here $g(\cdot;\theta)$ denotes the text encoder and $\theta$ is the collection of parameter of $g$, which is often initialized by BERT~\cite{bert}.  
The learning objective for dense retrieval can be expressed as

\begin{small}
\begin{equation}
\begin{aligned}
&\theta^{*} = \arg \min _{\theta} \ell(\theta)= \\
& \quad -\mathbb{E}_{q \sim p(\cdot)} \mathbb{E}_{d^{+}\sim p_{\text{pos}}(q)} \mathbb{E}_{d^{-} \sim p_{\text{neg}}(q)} \log p_{\theta}\left(d^{+}\mid q, d^{-}\right), 
\end{aligned}
\label{eq:loss}
\end{equation}
\end{small}

\noindent where $p(\cdot)$ is the distribution of queries, and $d^{+}$ and $d^{-}$ are sampled from the distribution of positive and negative document for $q$ (denoted as $p_{\text{pos}}(q)$ and $p_{\text{neg}}(q)$), respectively. In practice, the negative documents can either be BM25 negatives~\cite{dpr} or mined by DR models from the past episode~\cite{ance}.   

During training, we aim to maximize the probability  of selecting the ground-truth document $d^{+}$ over the negative document $d^{-}$ as 

\begin{small}
\begin{equation}
\begin{aligned}
\setlength{\abovedisplayskip}{1pt}
\setlength{\belowdisplayskip}{1pt}
p_{\theta}(d^{+}|q,& d^{-})=\frac{\exp{(f(q, d^{+}; \theta))}}{\exp{(f(q, d^{+}; \theta))}+ \exp{(f(q, d^{-}; \theta)})},
\end{aligned}
\label{eq:contra} 
\end{equation}
\end{small}

This dense retrieval configuration has shown strong empirical performances in a wide range of supervised scenarios, where the training and testing data are drawn from the same distributions, and a large amount of relevance labels are available~\citep{dpr, ance, rocketqa}.

\subsection{ZeroDR and Distribution Shifts}
\label{sec:motivation_shift}
Unlike supervised settings, the empirical advantages of dense retrieval are more ambivalent in zero-shot scenarios~\citep{beir}. We first discuss the common setups of ZeroDR and then investigate the impact of distribution shifts on zero-shot performance of dense retrieval models.

\textbf{ZeroDR Task.} A retrieval task is considered zero-shot if no task-specific signal is available. Unless in large commercialized scenarios like web search, zero-shot is often the norm, \emph{e.g.}, when building search systems for a new application,  in domains where annotations require specific expertise, or in personalized scenarios where each user has her own corpus.

Besides relevance labels, the availability of in-domain queries is also a rarity---often only a few example queries are available.
The most accessible in-domain information is the \textit{corpus}, which is a prerequisite to build search systems. Sparse retrieval needs to pre-build the inverted index before serving any query; dense retrieval systems have to pre-compute the document embeddings.

These properties of zero-shot retrieval lead to a common ZeroDR setup where  models can leverage the target corpus to perform unsupervised domain adaptation, but their supervised training signals only come from the source retrieval task, namely {\marco}~\citep{modir, wang2021gpl}.  

In this paper, we follow the standard practice in recent ZeroDR research, with MS MARCO passage retrieval~\citep{msmarco} as the source retrieval task, the tasks collected in the BEIR benchmark~\citep{beir} as the zero-shot target, and the corpora of BEIR tasks available at training time for unsupervised domain adaptation.

% While the labeled query-document relevancy scores can be expensive to obtain, unlabeled corpora from target tasks are often required to preprocessing and building indexes for both sparse and dense retrieval systems. 
% Thus, it is common for ZeroDR to leverage the corpus from target tasks during training (but without relevance labels)~\cite{genq,wang2021gpl,modir,thakur2022domain}.

% In this work, we use {\marco}~\cite{msmarco} as the source task, which is a common setting for ZeroDR methods~\cite{beir}. 
% Training on {\marco} also achieves superior performance over other alternatives (e.g. NQ) as demonstrated in recent studies~\cite{ren2022thorough}.

\textbf{Distribution Shifts.} Before discussing our ZeroDR method, we first study the distribution shifts between the source training task (MARCO) and the zero-shot target tasks (BEIR). 
%cx{only need half of space used currently}

Following the analysis in \citet{beir}, we use pairwise weighted Jaccard similarity~\cite{jaccard} to quantify the distribution shifts both at the query side and the document side.
The  document distribution shift  is measured directly at the lexicon level,
by the similarity of their unigram word distributions.
The query distribution shift is measured on the distribution of query types, using the nine-type categorization from~\citet{ren2022thorough} (more details in Appendix~\ref{app:query_type}).
As shown in \citep{ren2022thorough}, search intent types are more representative than lexicon for short queries.

Figure~\ref{fig:motivation-qdsim} plots the distribution shifts from MARCO to BEIR tasks and the corresponding performance differences between dense retrieval and sparse retrieval. We use BM25 as the sparse retrieval method and ANCE starting from pretrained BERT~\citep{ance} and coCondenser~\citep{cocondenser} as representative DR models. 

The average similarity between {\marco} and BEIR tasks are 32.4\% and 34.6\% for queries and documents, indicating the existence of  significant distribution shifts from MARCO to BEIR.
Furthermore, these shifts are correlated with the performance degradation of dense retrieval models, as DR models perform much worse than BM25 on BEIR tasks that are less similar to {\marco}.  
The contrastive learning on MARCO does not address this challenge; ANCE initialized from coCondenser still underperforms BM25 on BEIR tasks where distribution shifts are severe.

\section{{\model} Method}
\label{sec:method}
To combat the distribution shifts from training source to zero-shot targets,
\model{} introduces two training techniques: COntinuous COntrastive pretraining (COCO) and implicit Distributionally Robust optimization (iDRO).
The first \textit{continuously pretrains} the language model on target corpora to handle document distribution shifts.
The latter improves the model robustness during \textit{fine-tuning}, which then lead to better generalization for unseen target queries.
This section describes these two components in detail.

\subsection{Continuous Contrastive Pretraining}
\label{sec:coco}

Sequence Contrastive Learning (SCL) aims to improve the alignment of similar text sequences in the pretrained representations and the uniformity of unrelated text sequences~\cite{meng2021cocolm}, which benefits supervised dense retrieval~\cite{cocondenser, ma2022pre}.
In zero-shot settings, however, SCL-pretrained models still suffer from the distribution shifts, as observed in Figure~\ref{fig:motivation-qdsim}.

COCO addresses this challenge via continuously pretraining the language model on the target corpora, using the contrastive learning settings widely adopted in recent research~\citep{gtr, cocondenser,  cpt}. 

Specifically, for each document $d_i$ in target corpora, we randomly extract two disjoint sequences $s_{i,1}$ and $s_{i,2}$ from  $d_i$ to form the positive pair in:

\begin{small}
\begin{align}
\mathcal{L}_{\text{co}}&=\sum_{i=1}^{n}\ell(s_{i,1}, s_{i,2})) \label{eq:cl} \\
&=\sum_{i=1}^{n}-\log \frac{\exp (\langle g(s_{i,1}), g(s_{i,2})\rangle)}
{ \sum_{j=1,2} \sum_{s^- \in B} \exp (\langle g(s_{i,j}), g(s^-)\rangle)}. \nonumber
\end{align}
\end{small}
The contrastive loss with sequence representations $g(s)$ and in batch negatives $s^- \in B$. 

This contrastive learning is used in combination with language modeling~\citep{cocondenser} to continuous pretrain on target corpora~\citep{gururangan2020dont}. It adapts the language models to target corpora before fine-tuning on source labels, to reduce the impact of document distribution shifts.

\subsection{Distributionally Robust Optimization}
\label{sec:idro}

The query distribution shifts are more challenging, as often target queries are only available, if any, at a small amount.
For example, applying COCO on a few queries is unlikely useful.

To address this challenge, we exploit the assumption from distributional robust optimization (DRO): a model trained to be more robust on the source domain is likely to better generalize to unseen data~\citep{Sagawa2020Distributionally, wiles2021fine}. 
In addition, as explicit target domain/group information is unavailable, we perform implicit DRO (iDRO) to improve models' robustness regarding to source query clusters during fine-tuning.

\textbf{iDRO Loss.} Specifically, we first cluster source queries using K-Means~\citep{lloyd1982least} on their embedding similarities (dot-product) from COCO, and then optimize the following iDRO loss:
\begin{align}
&\mathcal{L}_{\text{iDRO}}(\theta) = \sum_{i=1}^{K}\alpha_{i} \omega_i\ell_i(\theta), \label{eq:idro} \\
 &\alpha_{i} \propto [\ell_i(\theta)]^{\beta}; \beta \geq 0.
\end{align}
It weights the per cluster dense retrieval loss  $\ell_i(\theta)$ in Eqn.~\ref{eq:loss} of $K$ total clusters using two parameters.
The first one, $\alpha_{i}$, up-weights clusters with higher training loss, with the emphasize on harder clusters defined by hyperparameter $\beta$.
The second one $\bm{\omega}\in \mathbb{R}^{K}$ is learned to maximize the loss decreases on all clusters, which we derive a closed form solution in the rest of this section.

\textbf{Dynamic Cluster Weighting.} An ideal choice of $\boldsymbol{\omega}^t$ at training step $t$ would provide biggest reduction on the training loss of all query clusters, but is difficult to obtain. To derive a closed form solution of $\omega^t$, we approximate the loss reduction using first order Taylor expansion:
\begin{align}
&\ell_{\text{g}}
= \sum_{i=1}^{K} \left(\ell_{i}(\theta-\eta \nabla_{\theta} \mathcal{L}_{\text{iDRO}}(\theta)) -\ell_{i}(\theta)\right) \label{eq:grad_step} \\ 
&\approx -\eta\sum_{i=1}^{K}\sum_{j=1}^{K}\alpha_{i}\alpha_{j}\omega^t_i\left(\nabla_{\theta}\ell_i(\theta)\right)^{\texttt{T}}\nabla_{\theta}\ell_j(\theta). \label{eq:app_grad_step}
\end{align}
 Eqn.~\ref{eq:grad_step} is the loss reduction on all clusters, after a stochastic gradient descent operation with step size $\eta$. Eqn.~\ref{eq:app_grad_step} is its first order expansion.

In addition, we avoid potential rapid change of cluster weights for optimization stability, by adding a KL divergence regularization between $\boldsymbol{\omega}$ at different steps. This leads to the following optimization target:
\begin{align}
&\min _{\omega^{(t)}} \ \ell_{\text{g}} + \tau \mathcal{D}_{\text{KL}}(\boldsymbol{\omega}^{(t)}||\boldsymbol{\omega}^{(t-1)}), \\
&\text{ s.t.} \quad \sum_{i=1}^{K}{\omega_i^{(t)}}=1. \label{eq:constrain_reg}
\end{align}
The strength of KL regularization is controlled by hyperparameter $\tau$. 
By using Lagrangian multiplier (details in Appendix~\ref{app:idro}), the optimal weight for each group $\omega_i^{t*}$ can be  calculated as  
\begin{align}
&\omega_i^{t*} = \frac{\omega^{(t-1)}_i \exp\left(\frac{1}{\tau}\sum_{j=1}^{K}r_{ij}\right)}{\sum_{i=1}^{K}\omega^{(t-1)}_i \exp\left(\frac{1}{\tau}\sum_{j=1}^{K}r_{ij}\right)};  \\
&r_{ij}=[\ell_i(\theta)\ell_j(\theta)]^{\beta}\left(\nabla_{\theta}\ell_i(\theta)\right)^{\texttt{T}}\nabla_{\theta}\ell_j(\theta).
\label{eq:final_omega}
\end{align}
Intuitively, the optimal solution considers the gradient and loss similarity between different groups $r_{ij}$. It favors clusters sharing more `common needs'~\citep{cgd} with others to improve the model robustness across all clusters. 

COCO and iDRO operate at different training stages of dense retrieval.
COCO continuously pretrains the language model to adapt to the target documents, while iDRO improves the robustness of dense retrieval in the fine-tuning stage for better generalization on unseen queries.
The two together forms \model{} that aims to improve zero-shot retrieval accuracy by combating the distribution shift from both the query and the document side.

% To summarize, both COCO and iDRO are designed to tackle the distribution shift, but with different focuses. 
% COCO leverages the corpora from target tasks to improve the representations of text sequences on target domains, whereas iDRO leverages the training signals from source domain only and enhances the DR model's generalization ability over unseen queries during finetuning.   
% When combined together, we combat the distribution shift issue from both the query and document side in two training stages 
% and improve robustness of DR models to target tasks.

\section{Experiments}
\label{sec:exp}
In this section, we first describe our experiment setups and evaluate \model{}. Then we analyze the efficacy of {COCO} and iDRO.

\subsection{Experimental Setups}
Our experiments use the tasks collected in BEIR~\cite{beir}, a recent standard benchmark for zero-shot dense retrieval. The dataset details are in Appendix~\ref{appx:datasets}. 

\textbf{Baselines.} We consider various baselines, including standard sparse and dense retrieval models on BEIR.
We also follow~\cite{wang2021gpl} to further compare {\model} with dedicated ZeroDR approaches based on unsupervised domain adaptation: 
these models are first pretrained on the target corpus and then fine-tuned on {\marco}.
We list the details of baselines in Appendix \ref{appx:baseline}.

\begin{table*}[t]
  \small
  \centering
  \resizebox{\linewidth}{!}{%
    \begin{tabular}{@{}l|l|lllllllll|l|ll@{}}
\toprule
%  {{\textbf{Method} ($\rightarrow$)}}
 &  {\textbf{Sparse}}                & \multicolumn{9}{c|}{\textbf{Dense}}    &  {\textbf{Late-Inter.}}           & \multicolumn{2}{c}{\textbf{\model{} (Ours)}}     \\  \cline{2-14} 
 &  {BM25}
&  {DPR}   &  {ANCE}  
%  {TAS-B$^\#$}      
&  {Contriever} & {GenQ$^\dagger$} 
&  {GPL$^{\dagger,\#}$}  &  
 {GTR{\tiny{XL}}$^\ddagger$}         &  {GTR{\tiny{XXL}}$^\ddagger$} &
 {CPT{\tiny{L}}$^{\ddagger,\sharp}$}  &  {CPT{\tiny{XL}}$^{\ddagger,\sharp}$}  & {ColBERT}        &   {\text{Base}} & \multicolumn{1}{c}{\text{Large}}     \\ \hline
 \textbf{Parameters\#} &  {---} &  110M & 110M  & 110M& 66M*18 & 66M*18 & 1.2B & 4.8B & 6B & 175B  & 110M &110M & 335M \\ \hline
MS MARCO                &  {0.228}       
&  {0.354} &  {0.388}     &  {0.407} &  {0.408} 
&  {---}  &  {0.439}          &  {0.442} &  {---} &  {---}
    &  {0.401}        &  {0.419} & {0.424} \\
\hline
TREC-COVID                                           &  {0.656}          %& \textbf{0.713} 
&  {0.575} &  {0.654}
% &  {0.481}       
% &   {0.492} 
&  {0.596} &  {0.619} &   {0.700} &  {0.584} &  {0.501} &  {0.642} &  {0.649}  &  {0.677}  &  {\underline{0.789}} &    \textbf{0.804}  \\
BioASQ                                                &  \underline{0.465}          
%& 0.431          
&  {0.232} &  {0.306} 
% &  {0.383}       
% &  {0.308} 
&     {---}     
&  {0.398} &     {0.442}
&  {0.317}          &  {0.324}  
&  {---} &     {---}
  & \textbf{0.474}  &  {0.429} & {0.449}  \\
NFCorpus              &  {0.325}          %& 0.328        
&  {0.210} &  {0.237} 
% &  {0.319}       
 &  {0.328}   
&  {0.319} &  {0.345} &  {0.343} &  {0.342} &  {0.380}&  {\textbf{0.407}}     &  {0.305} &  \underline{0.355} & 0.354         \\
NQ          &  {0.329}          &  {0.398}          &  {0.446}   &  {0.498}  
&  {0.358} &  {0.483}  
&  {0.559$^*$}          &  {\textbf{0.568}$^*$}
&  {---} &  {---}             
&  {0.524} &  \underline{0.505} & \textbf{0.547} \\
HotpotQA                       &  {0.603} &  {0.371}           &  {0.456}  &  \underline{0.638}       &  {0.534} &  {0.582}      &  {0.591}          &  {0.599}    &  {0.648}    &  {\textbf{0.688}}      & 0.593 &  {0.616}  &  0.641       \\
FiQA-2018                                             &  {0.236}          &  {0.274}          &  {0.295}  &  {0.329}       &  {0.308} &  \underline{0.344}         &  {0.444}          &  {0.467} &  {0.452}    &  {\textbf{0.512}}      & 0.317 &  {0.307} & 0.329 \\
Signal-1M                                             &  {\textbf{0.330}} &  {0.238}          &  {0.249}  &  {---}       &  \underline{0.281} &  {0.276}        &  {0.268}          &  {0.273}    &  {---}    &  {---}      & 0.274 &  {0.271} & 0.285       \\
TREC-NEWS                                             &  {0.398}          &  {0.366}  &  {0.382}  &  {---}       &  {0.396} &  {\underline{0.421}}  &  {0.350}          &  {0.346}        &  {---}    &  {---}        & 0.393 &  {0.403} & \textbf{0.432}       \\
Robust04                                              &  {0.408}          &  {0.344}          &  {0.392}  &  {---}       &  {0.362} &  {0.437}        &  {0.479}          &  {\textbf{0.506}}    &  {---}    &  {---}    & 0.391 &  \underline{0.443} & {0.482} \\
ArguAna                                               &  {0.414}          &  {0.414}          &  {0.415}  &  {0.446}       &  {0.493} &  {\textbf{0.557}}        &  {0.531}          &  {0.540}     &  {0.469}    &  {0.435}   & 0.233 &  {0.493} & 0.515 \\
Touché-2020                                           &  {\textbf{0.367}} &  {0.208}          &  {0.240}  &  {0.230}       &  {0.182} &  \underline{0.255}        &  {0.230}          &  {0.256}  &  {0.309}    &  {0.291}     & 0.202 &  {0.238} & 0.263         \\
Quora                                                 &  {0.789}          &  {0.842}          &  {0.852}  &  {0.865}       &  {0.830} &  {0.836}      &  {0.890}          &  {\textbf{0.892}}     &  {0.677}    &  {0.638}     & 0.854 &  \underline{0.867} & 0.872\\
DBPedia-entity                                        &  {0.313}          &  {0.236}          &  {0.281}  &  \underline{0.413}       &  {0.328} &  {0.384}   &  {0.396}          &  {0.408}       &  {0.412}    &  {\textbf{0.432}}      & 0.392 &  {0.391} & 0.407 \\
SCIDOCS                                               &  {0.158}          &  {0.107} &  {0.122}  &  \underline{0.165}       &  {0.143} &  {{0.169}}          &  {0.159}          &  {0.161}    &  {---}    &  {---}    & 0.145 &  {0.160} & \textbf{0.178}       \\
Fever                                                 &  {0.753}          &  {0.589}          &  {0.669}  &  {0.758}       &  {0.669} &  \underline{0.759}   &  {0.717}          &  {0.740} &  {0.756}    &  {{0.775}}    & 0.771 &  {0.751}   & \textbf{0.793}      \\
Climate-Fever                                         &  {0.213}          &  {0.176}          &  {0.198}  &  \underline{0.237}       &  {0.175} &  {0.235}       &  {\textbf{0.270}} &  {0.267}      &  {0.194}    &  {0.223}   & 0.184 &  {0.211}   & 0.247      \\
SciFact                                               &  {0.665}          &  {0.475} &  {0.507}  &  {0.677}       &  {0.644} &  {0.674}           &  {0.635}          &  {0.662} &  {0.744}    &  {\textbf{0.754}}      & 0.671 &  \underline{0.709}  & 0.722      \\
CQADupStack                                           &  {0.299}          &  {0.281}          &  {0.296}  &  {0.345}       &  {0.347} &  {0.357}      &  {0.388}          &  {\textbf{0.399}}    &  {---}    &  {---}      & 0.350 &  \underline{0.370} & 0.393 \\ \hline
\bf Avg  CPT Sub                                                  &  {0.484}          &  {0.397}          &  {0.437}  &  {0.502}       &  {0.464} &  {0.516}       &  {0.511}          &  {0.516}    &  {0.517}    &  {{0.528}}     & 0.473 &  \underline{0.521} & \textbf{0.541}\\
\bf Avg                                                   &  {0.428}          &  {0.352}          &  {0.389}  &  {---}       &  {0.410} &  {0.459}        &  {0.453}          &  {0.458}    &  {---}    &  {---}    & 0.431 &  \underline{0.462}& \textbf{0.484} \\
% Avg w/o MS Marco                                      &  {0.423}          & {\underline{0.434}}    &  {0.237} &  {0.389} &  {{\underline{0.415}}} &  {0.410} & 0.431          &  {0.416}    &  {0.445}     &  {0.453}          & \textbf{0.458} \\ 
\bottomrule
\end{tabular}
  }
  \caption{nDCG@10 on the BEIR benchmark. The best result for each task is marked \textbf{bold}, and the best result among \emph{fair} baselines (using BERT-base or smaller models as the backbone) is \underline{underlined}. 
  Avg CPT Sub is the average performance on 11 BEIR tasks used in \citet{cpt}.
 $^*$: Unfair comparison, NQ is used in training for GTR. 
 $\dagger$: Train an independent model for each task.\
  $\ddagger$: Larger Model, more training data.   
  $\#$: Use cross-encoders reranking teachers.
  $\sharp$: Can only be accessed with paid APIs.}
%   Single model/vector, dense retrieval}
  \label{tab:all_ndcg}
\end{table*}
% \vspace{-1ex}

\textbf{Implementation Details.}  
For \model{}, we use the same architecture as BERT~\cite{bert} and consider both \emph{Base} and \emph{Large} size in our experiments. 
The architecture of {\cocobase} is the same as BERT$_{\text{Base}}$: 
12 layer Transformer, 768 hidden size. Similarly, the architecture of {\cocolarge} model is the same
as BERT$_{\text{Large}}$, using 24 layer and 1024 hidden size.  
Our implementation uses PyTorch~\cite{paszke2019pytorch} with Hugging Face Transformers~\cite{transformers} and OpenMatch~\cite{openmatch} codebase.

In COCO stage, we initialize our model with Condenser~\cite{gao-callan-2021-condenser}, and continuously pretrain the model for 8 epochs (around 200K steps) on the corpus of BEIR and {\marco}. 
We optimize the model using AdamW~\cite{loshchilov2018adamw} with a peak learning rate 1e-4/1e-5 for Base/Large, weight decay 0.01, and linear learning rate decay. 
The model is trained with 8 Nvidia A100 80GB GPUs and FP16 mixed-precision training. The batch size for each GPU is set to 200. Maximum number of tokens per sequence is 128. 

The iDRO stage trains on MARCO passage retrieval with AdamW, 5e-6 learning rate, linear learning rate schedule, and batch size 64 for each GPU.
Following \citet{ance}, the model is first trained using BM25 negatives and then on self-negatives from the DR model. 
We update the query clusters with K-Means ($K=50$) when refreshing negative samples. 
The running time for COCO and iDRO are around 1.5 days each for {\cocobase} and around 3 days for {\cocolarge}.

\textbf{Evaluation Details.} When evaluating on the BEIR benchmark, we use
sequences of 64 tokens for the questions and 128 for the documents in all datasets except TREC-NEWS, Robust04, SciFact and ArguAna. In particular, we set the document length to 256 for TREC-NEWS, Robust04 and SciFact as they have larger document length on average. For ArguAna, we set both question and document length to 128 as it has longer queries. 

\textbf{Hyperparameters.} The main hyperparameters in {\model} includes the number of groups $K$, the temperature parameter $\tau$ and  the importance factor $\beta$. We keep $\beta=0.25$ in {\model} and study the effect of $N$ and $\tau$ in Sec.~\ref{sec:abla}.

% \section{Evaluation Results}

\subsection{Overall Results}

\begin{table*}[!t]
%   \small
% \vspace{-1.5ex}
  \centering
  \resizebox{0.9\linewidth}{!}{%
    \begin{tabular}{l|lll|lll|lll|cll|ll}
\toprule
\bf Method ($\rightarrow$) & \multicolumn{3}{c|}{\bf COCO-DR Base}  & \multicolumn{3}{c|}{\bf COCO-DR Large}      & \multicolumn{3}{c|}{\bf coCondenser}   &
\multicolumn{2}{c}{\bf Condenser}  \\ 
\bf Dataset ($\downarrow$) & Full & -iDRO & -COCO & Full & -iDRO & -COCO & Base \shortcite{cocondenser} & Base & Large & Base & Large  \\ \hline
TREC-COVID     &
\textbf{0.789} &	0.771 &	0.763 &	\bf 0.804&	0.797&	0.745& 0.715 & 0.758 & 0.745 & 		0.728 & 0.780 \\
BioASQ      & 
\textbf{0.429} &	0.424 &	0.353 &	0.449&\bf 	0.450&	0.413& 	0.318 & 0.341 & 0.410 & 	0.330 & 0.381 \\
NFCorpus             & 
\textbf{0.355} &	0.354 &	0.333 &	\bf 0.354&	0.353&	0.349& 	0.307 & 0.326 & 0.350 & 	0.282 & 0.317 \\
NQ          & 
0.505 &	0.503 &	\textbf{0.506} & \bf 0.547&	0.536&	0.519 &	0.494 &	0.503 & 0.516 & 	0.472 & 0.492\\
HotpotQA         &
\textbf{0.616} &	0.610 &	0.592 &	0.641&\bf 	0.644&	0.614 & 	0.566 &0.584 & 0.616& 	0.572 & 0.591 \\
FiQA-2018    & 
0.307 &	0.302 &	\textbf{0.312} & \bf  0.329&	0.322&	0.328&	0.285 &0.303 & 0.326 & 		0.254 & 0.280 \\
Signal-1M          & 
0.271 &	0.275 &	\textbf{0.281} &0.285&	0.285	&\bf 0.296	&		0.274 &	0.274 & 0.295 &0.266 & 0.284\\
TREC-NEWS               &
0.403 &	0.398 &	\textbf{0.426} & \bf 0.432	&0.426&	0.413&	0.389 &	0.400 & 0.416 &		0.375 & 0.423 \\
Robust04      &
0.443 &	0.443 &	\textbf{0.446} &\bf 0.482&	0.467&	0.466&	0.399 &	0.442 & 0.461 & 		0.385 & 0.418 \\
ArguAna          & 
\textbf{0.493} &	0.479 &	0.473 &	  \bf 0.515&	0.513&	0.488& 	0.411 &	0.460 &0.484 & 0.439 & 0.469\\
Touché-2020      & 
0.238 &	0.238 &	\textbf{0.257} & \bf 0.263	&0.258	&0.249&	0.190 &	0.240 & 0.246 & 	0.236 & 0.244\\
Quora                & 
0.867 &	\textbf{0.868} &	0.862 &\bf 0.872&	0.869&	0.865&		0.863 &	0.860 & 0.862 & 0.855 & 0.852 \\
DBPedia-entity              & 
\textbf{0.391} &	0.389 &	0.382 &\bf 0.407&	0.401&	0.388&	0.356 &	0.364 & 0.386 &	0.362 & 0.364 \\
SCIDOCS            & 
0.160 &	\textbf{0.161} &	0.154 &\bf 0.178&	0.176&	0.171&	0.140 &0.150 & 0.171 & 		0.143 & 0.161 \\
Fever            & 
0.751 &	\textbf{0.757} &	0.739 &\bf 0.793&	0.783&	0.741&		0.678 &	0.751 & 0.724 & 0.725 & 0.736 \\
Climate-Fever         & 
\textbf{0.211} &	0.209 &	0.202 &\bf 0.247&	0.240&	0.233&	0.184 &0.208 & 0.226 & 		0.206 & 0.216 \\
SciFact               & 
\textbf{0.709} &	0.688 &	0.615 &\bf 0.722&	0.709&	0.696&	0.600 & 0.602 & 0.686 &		0.581 & 0.661 \\
CQADupStack         & 
\textbf{0.370} &	0.365 &	0.349 &\bf 0.393	&0.385	&0.367&		0.330 & 0.342 & 0.363 &	0.313 & 0.343 \\ \hline
Avg           & 
\textbf{0.462$^{\dagger,\ddagger,\flat,\natural}$} &	0.457 &	0.447 &	\bf 0.484$^{\dagger,\ddagger,\flat,\natural}$ &	0.478 &	0.463 & 	0.417 & 0.440 & 0.460 &	0.418 & 0.445 \\
% Avg w/o MS Marco                                      & \multicolumn{1}{c|}{0.423}          & {\underline{0.434}}    & \multicolumn{1}{c|}{0.237} & \multicolumn{1}{c|}{0.389} & \multicolumn{1}{c|}{{\underline{0.415}}} & \multicolumn{1}{c|}{0.410} & 0.431          & \multicolumn{1}{c|}{0.416}    & \multicolumn{1}{c|}{0.445}     & \multicolumn{1}{c|}{0.453}          & \textbf{0.458} \\ 
\bottomrule
\end{tabular}
  }
  \caption{Ablation study of \model{} without iDRO (-iDRO) or continuous contrastive (-COCO). 
  Apart from \shortcite{cocondenser}, all the results are based on our own implementations.
  Superscripts indicate statistically significant results with $p$-value $<0.01$ over -iDRO$\text{}^{\dagger}$, -COCO$\text{}^{\ddagger}$, coCondenser$\text{}^{\flat}$, Condenser$\text{}^{\natural}$.
%   The results of ``-COCO” only pretrain and finetune on {\marco}, similar to {\cocondenser}. 
  }
  \label{tab:abla}
\end{table*}

Table~\ref{tab:all_ndcg} shows the results on BEIR. 
Due to space limits, we only present the strongest baselines---other reported numbers are directly comparable, if they follow the standard ZeroDR settings on BEIR.

{\cocobase} outperforms all previous methods on the average retrieval accuracy of all BEIR tasks, with large margin improvements over previous systems at BERT$_\text{Base}$ scale.
It is also competitive and often better than models with significantly more parameters.
{\cocobase} achieves better average performance than GTR{\tiny{XXL}} and CPT{\tiny{L}} despite only using around 2\% of their parameters. 
With more parameters, {\cocolarge} outperforms the giant CPT{\tiny{XL}} model (175B) by 2.5\%, when evaluated on a subset of 11 datasets used in their experiment.
It is worth noting that CPT{\tiny{XL}} can only be accessed with paid APIs. One inference for 18 BEIR tasks costs around 1.4 million dollars\footnote{The embedding model price (\$0.2 per 1k tokens) at \url{https://openai.com/api/pricing} as of Oct. 2022.}. 
Scaling up models is not the only solution for  zero-shot capacity. Better methodologies to tackle the distribution shifts can also improve the generalization of dense retrieval models, while being much ``greener''~\cite{green}.

{\model} also outperforms GPL, the strong domain adaptation model for ZeroDR~\citep{wang2021gpl}.
Note that GPL leverages a query generation model to produce pseudo relevance labels for each BEIR task, uses a cross-encoder to filter the pseudo labels, and trains one retrieval model for each task.
\model{} does not rely on any of these techniques and uses one single model for all tasks. Its only modifications are on the model pretraining and fine-tuning strategies. More detailed comparisons with other domain adaptation approaches are in Sec.~\ref{sec:coco_exp}.

\begin{figure}[t]
    % \vspace{-4mm}
        \centering
        \hspace{-4mm}
        \subfigure[Effect of $K$]{
    \includegraphics[width=0.51\textwidth]{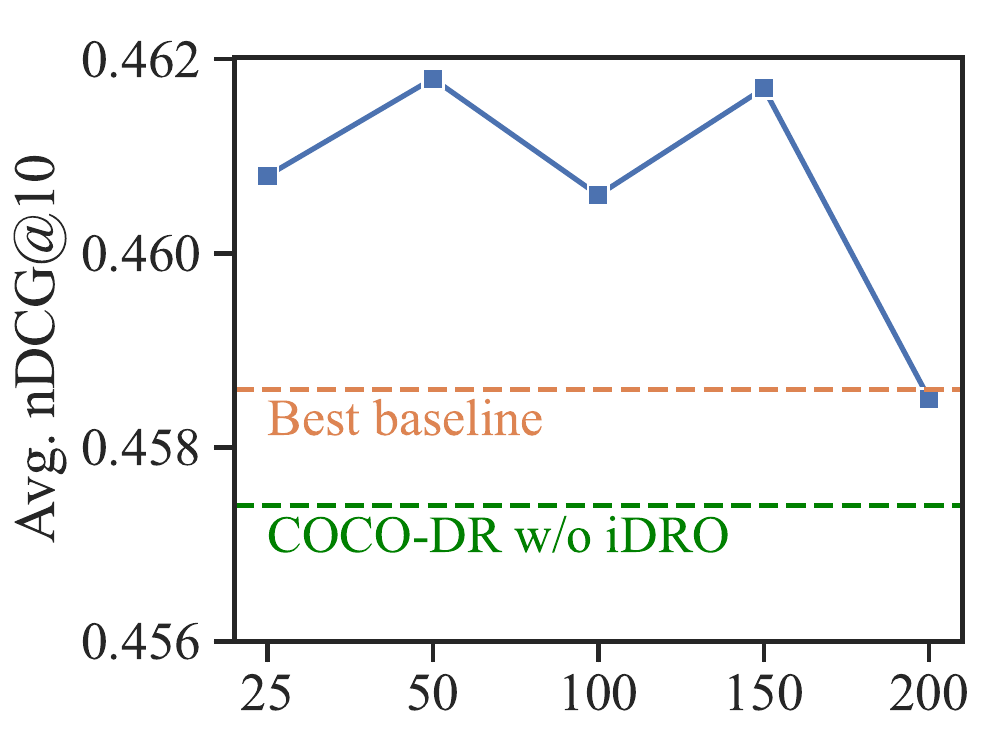}
            \label{fig:ablation_badge}
        }\hfill  \hspace{-7mm}
        \subfigure[Effect of $\tau$]{
            \includegraphics[width=0.51\textwidth]{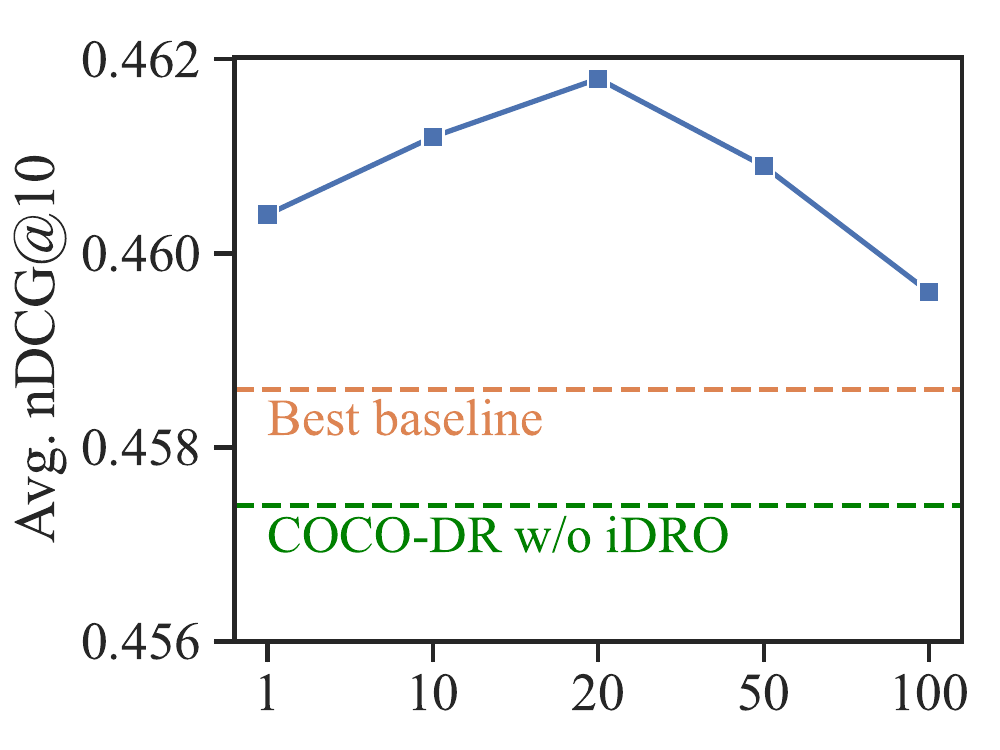}
            \label{fig:ablation_components}
        }
        \caption{
        Average NDCG@10 on BEIR of {\model} with different hyperparameters. The best baseline is GPL according to table \ref{tab:all_ndcg}.
        } \label{fig:param}
        \vspace{-1ex}
\end{figure}

\subsection{Ablation Study}
\label{sec:abla}

We perform two groups of ablations on \model{}'s  hyperparameters and components.

  \textbf{Hyperparameters.}   Figure~\ref{fig:param} shows the effect of two main hyperparameters, $K$ for K-Means clustering and $\tau$ for temperatures in iDRO. 
 When $K$ becomes very large, the performance decreases as there exist fragmented clusters that are not close to any target BEIR tasks. As a result, focusing on these clusters hurts the average performance on BEIR tasks. 
 When $\tau$ is too big, the weight for each group will be the same. On the contrary, if $\tau$ is too small, the model focuses too much on a few specific groups. 
 Nevertheless, iDRO is robust and outperforms the best baseline in most studied hyperparameter regions. 
\begin{figure}[t]
\centering
\vspace{-1ex}
\hfill 
\subfigure[TREC-COVID]{
    \includegraphics[width=0.488\textwidth]{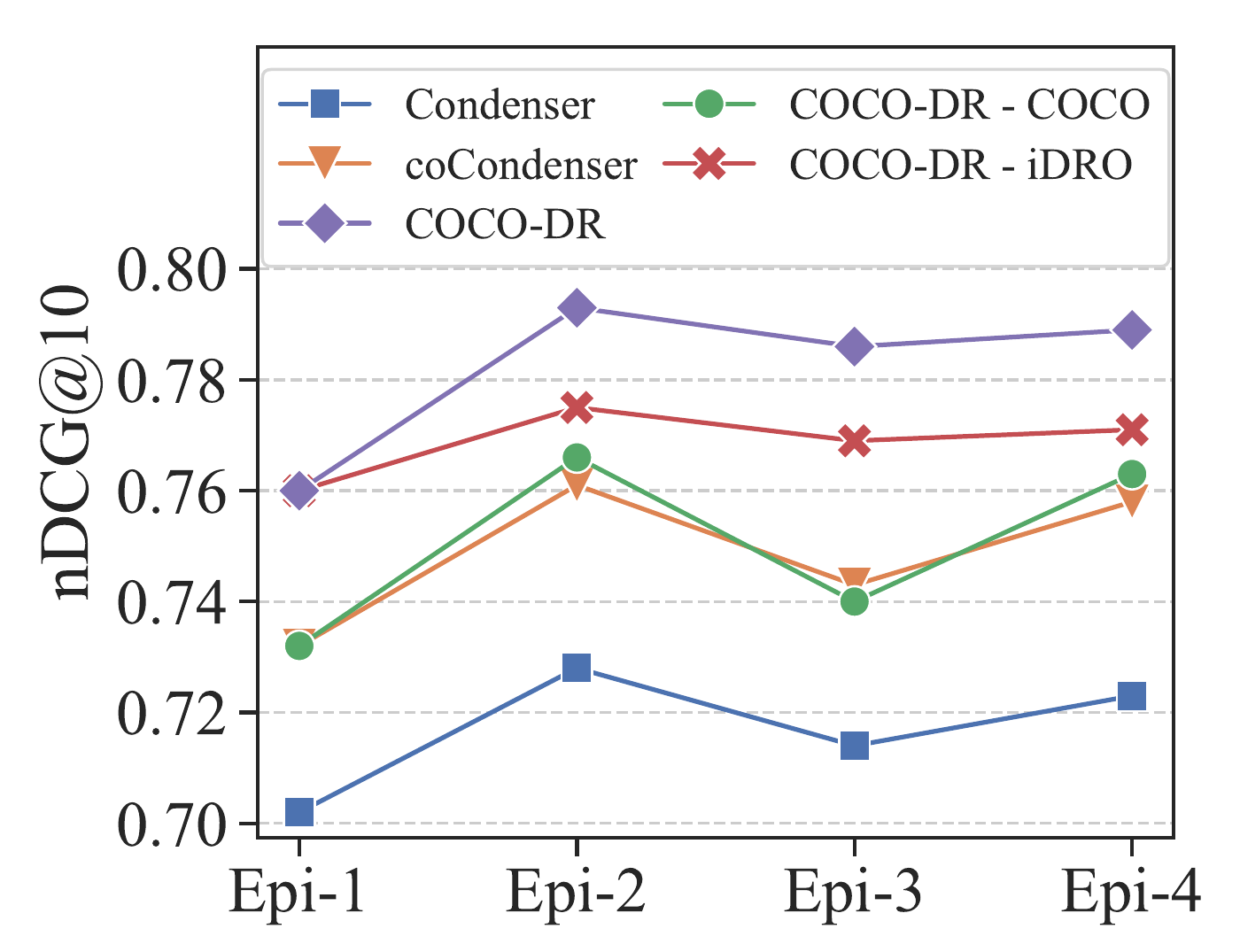}
            \label{fig:ablation_treccovid}
        }\hfill   \hspace{-3mm}
    %  \caption{}
% \subfigure[BioASQ]{
%     \includegraphics[width=0.331\textwidth]{Figures/abla_bioasq.pdf}
%             \label{fig:ablation_bioasq}
%         }\hfill   \hspace{-6mm}
    %  \caption{BioASQ}
\subfigure[SciFact]{
    \includegraphics[width=0.488\textwidth]{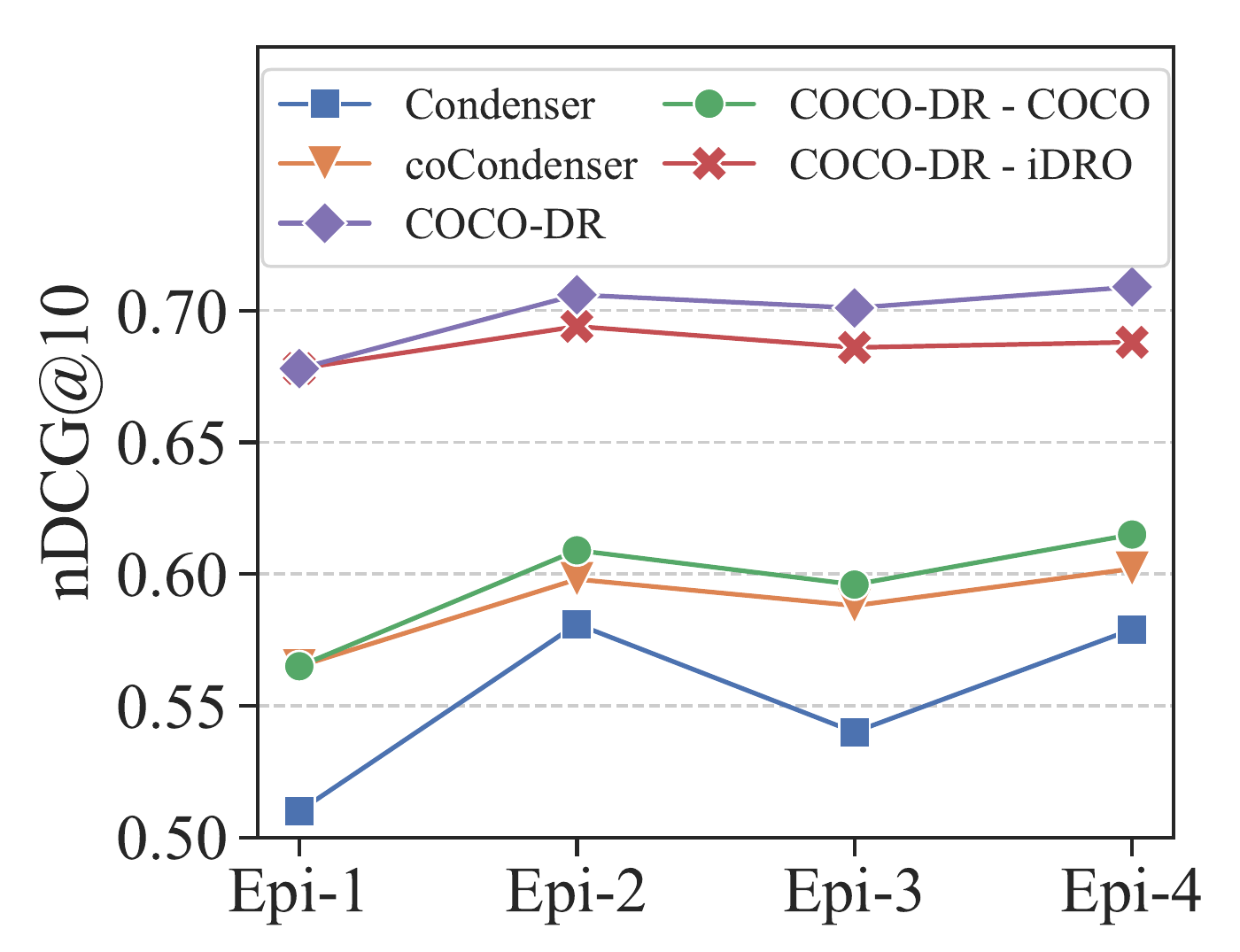}
            \label{fig:ablation_scifact}
        }\hfill 
        \vspace{-1ex}
\caption{The performance of {\model} and its variants over different training stages on TREC-COVID and SciFact. Epi-1 stands for the result after BM25 warmup, and Epi-2,3,4 are results of training with self-negative (ANCE). 
More results are in Appendix~\ref{app:dynamics}.
\vspace{-1ex}
}
\label{fig:beir_dynamics_main}
\end{figure}

\begin{figure*}[!t]
    % \vspace{-4mm}
        \centering
        \hspace{-2mm}
        \subfigure[COCO loss v.s. BEIR gain]{
    \includegraphics[width=0.33\textwidth]{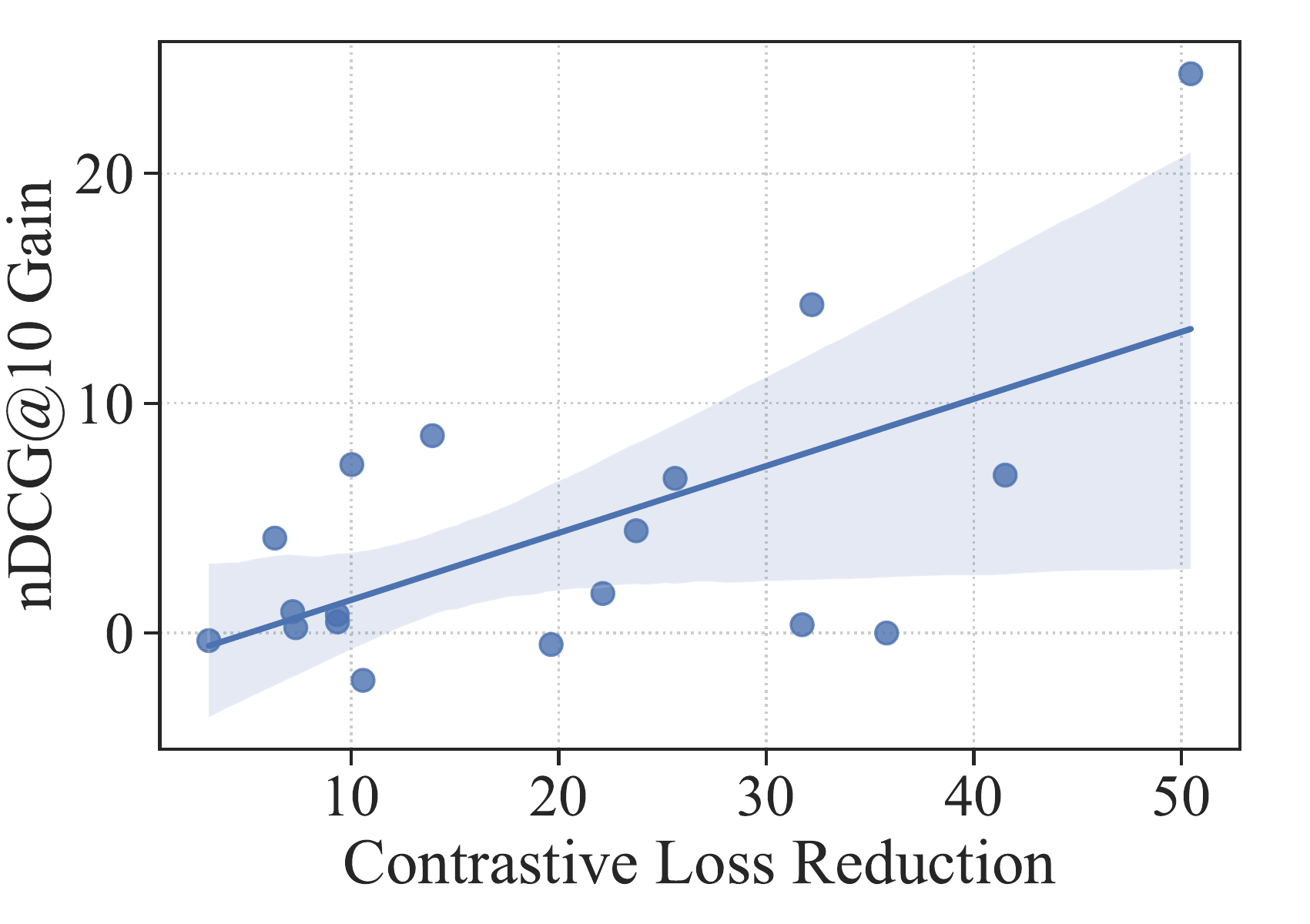}
            \label{fig:coco_gain}
        }\hfill  \hspace{-4mm}
        \subfigure[$\ell_{\text{uniform}}$ and $\ell_{\text{align}}$]{
            \includegraphics[width=0.335\textwidth]{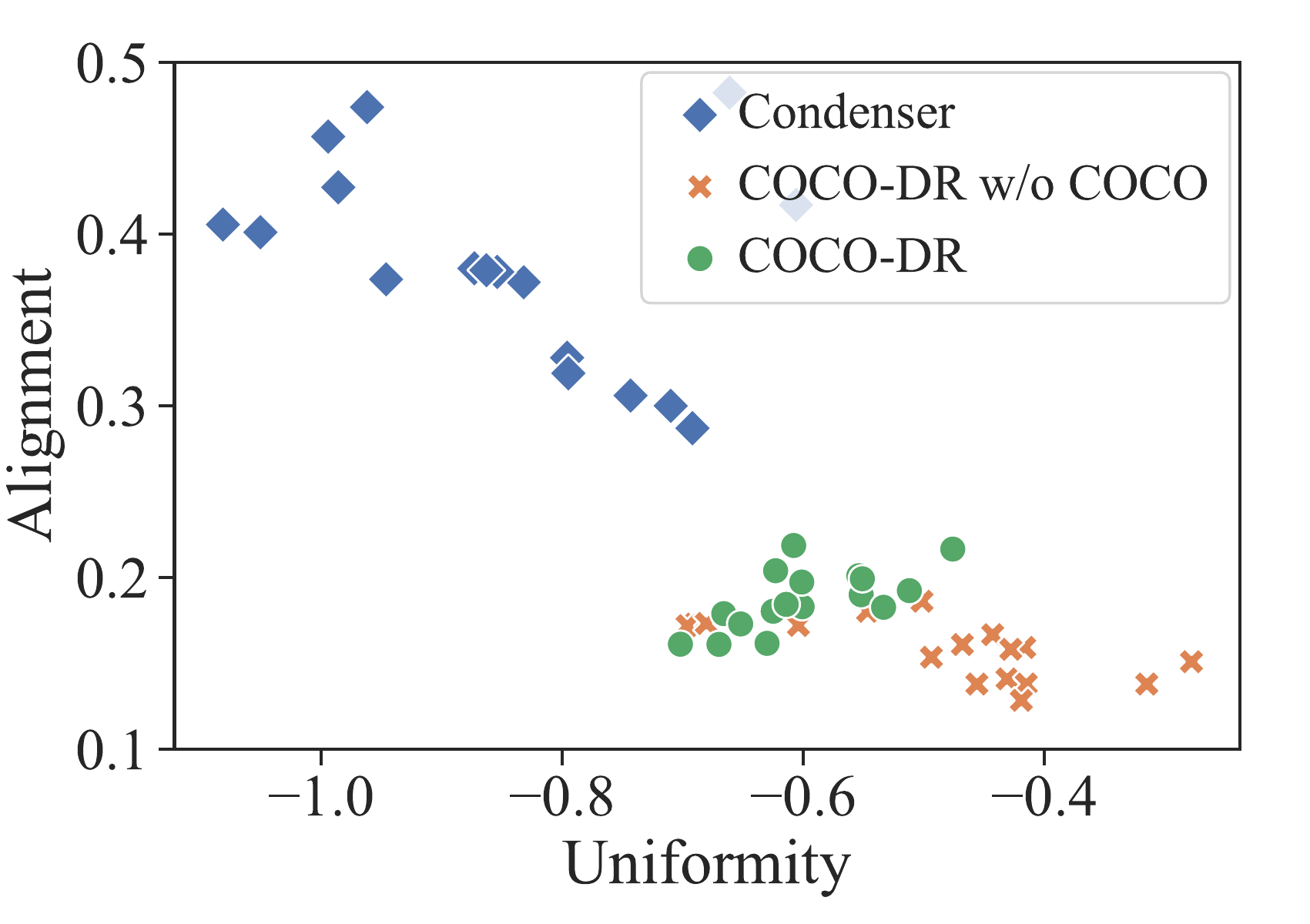}
            \label{fig:uni_align}
        }\hfill  \hspace{-4mm}
        \subfigure[BEIR v.s. {\marco} nearest groups]{
            \includegraphics[width=0.33\textwidth]{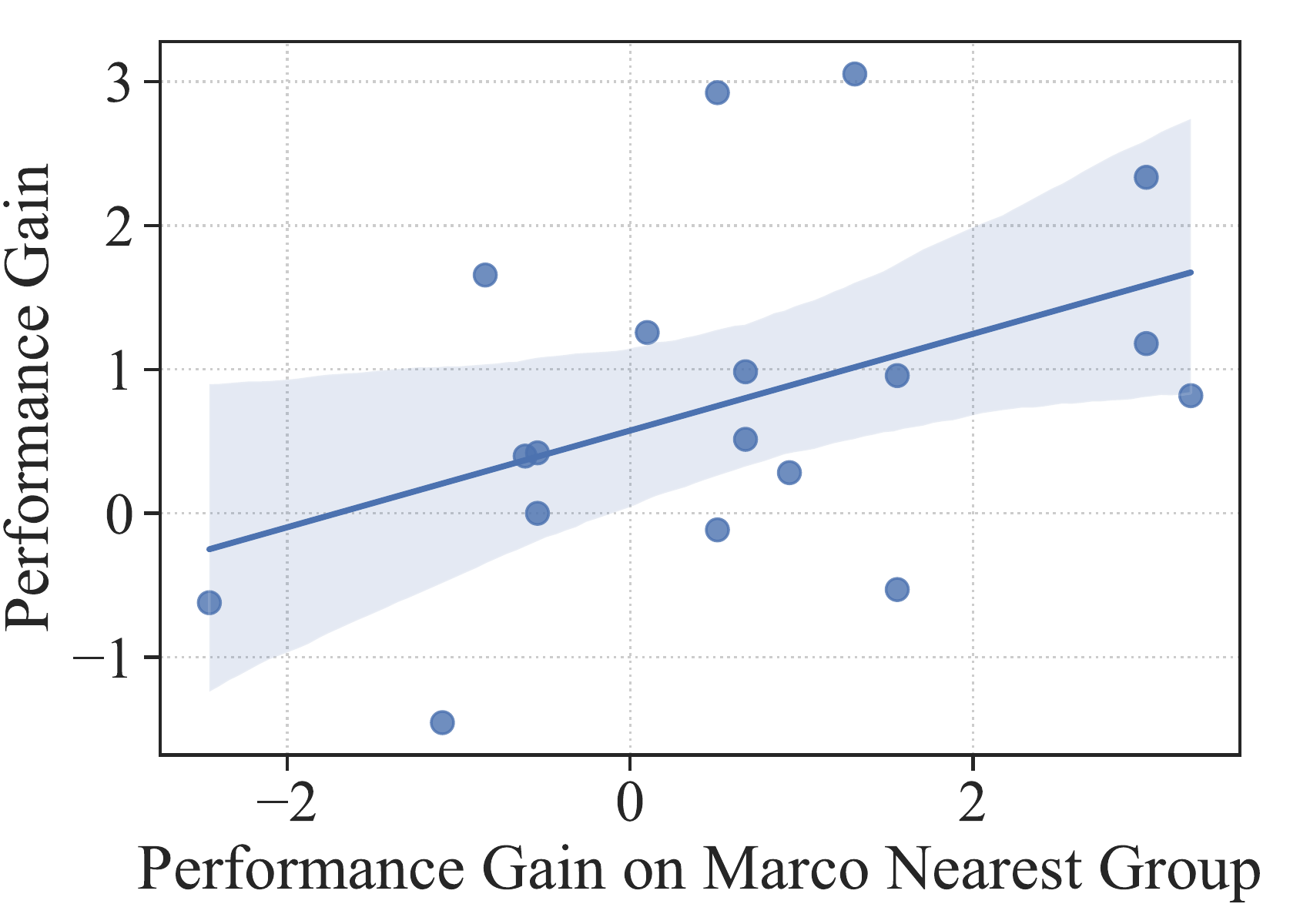}
            \label{fig:idro_gain}
        }
        % \vspace{-0ex}
        \caption{
        Left: The relation between the gain of COCO v.s. the gain on BEIR  tasks.
        Middle: $\ell_{\text{uniform}}$ \& $\ell_{\text{align}}$ plot for {\model} and its variants on BEIR tasks.
        Right: The relation between the gain on BEIR tasks v.s. the gain on nearest {\marco} groups. 
        } \label{fig:coco_detail}
\end{figure*}

 \textbf{Designed Components.} Table~\ref{tab:abla} shows the performance of \model{} variations and the pretraining baselines.
COCO and iDRO improve the average performance on BEIR datasets by 3.9\% and 1.1\% relatively. 
The stronger relative gains from COCO is expected, as it leverages the available in-domain corpora, while iDRO is designed for a harder challenge: to improve model generalization ability w.r.t. unseen target queries solely using training signals from the source.

Compared with {\cocondenser} which is pretrained on {\marco} only (-COCO) and uses the standard DR loss during finetuning (-iDRO), each design individually leads to improvements over a majority of (COCO on 16; iDRO on 14) the 18 tasks included in BEIR.
These two focus on different distribution shifts and operate at different stages of the training pipeline. Combining them in COCO-DR provides the best overall effectiveness. 

Figure~\ref{fig:beir_dynamics_main} zooms in the performances of COCO-DR and its variations on two BEIR tasks, TREC-COVID and SciFact, at different fine-tuning stages on the source task. 
It shows that COCO also helps stabilize the fine-tuning step on {\marco} and reduces the oscillation between different training iterations.
The benefit of iDRO is strong on biomedical tasks as shown in Figure~\ref{fig:beir_dynamics_main}, as {\marco} indeed has relevent search intents in the BioMed domain. 
In Section \ref{sec:coco_exp} and \ref{sec:idro_exp}, we analyze the benefits of the two designs in detail.

\subsection{Influence of COCO Pretraining}

% 0.233	0.336		0.571	0.246	0.263	0.3298
% 0.296	0.502		0.66	0.297	----	
% 0.267	0.55		0.683	0.29	0.379	0.4338
% 0.27	0.585		0.697	0.313	0.374	0.4478
% 0.302	0.6		0.695	0.304	0.388	0.4578
% 0.293	0.628		0.761	0.318	0.394	0.4788
% 0.27	0.627		0.654	0.306	0.345	0.4404
% 						#DIV/0!
% 						#DIV/0!
% 0.287	0.638		0.724	0.33	0.381	0.472
% 0.328	0.664		0.726	0.345	0.414	0.4954
% 						#DIV/0!
% 0.306	0.706		0.807	0.372	0.44	0.5262
% 0.331	0.676		0.712	0.368	0.467	0.5108
% 0.364	0.683		0.714	0.381	0.483	0.525

\begin{table}[t!]
\small
\centering
\resizebox{1\columnwidth}{!}{
		\begin{tabular}{l@{\hskip2pt}|@{\hskip2pt} l@{\hskip2pt}|@{\hskip2pt} l@{\hskip2pt}|@{\hskip2pt} l@{\hskip2pt}|@{\hskip2pt}l@{\hskip2pt}|@{\hskip2pt}l@{\hskip2pt}|@{\hskip2pt}l@{\hskip2pt}}
		   \toprule
		    & \bf \multirow{2}{*}{FiQA} & \bf \multirow{2}{*}{SciFact} & \bf TREC- & \bf CQAD- & \bf \multirow{2}{*}{Robust04} & \bf \multirow{2}{*}{Avg.} \\
		    \bf Model  & & & \bf Covidv2 & \bf upStack &  \\
		   \hline 
		   \multicolumn{7}{l}{\bf Sparse Retrieval} \\ %& \multicolumn{2}{c|}{80.0}  & -\!-\!- \\
		   \hline
		   BM25~\shortcite{bm25}  & 0.239 & 0.661 & 0.601 & 0.315 & 0.387 & 0.461 \\
		   \hline 
		  % \midrule
		   \multicolumn{7}{l}{\textbf{Domain Adaptation Methods}}\\
		   \hline 
		   UDALM~\shortcite{udalm} & 0.233 & 0.336 & 0.571 & 0.246 & 0.263 & 0.330 \\
		   MoDIR~\shortcite{modir}  & 0.296	& 0.502	&	0.660 &	0.297 & --- & --- \\
		  % \hline 
		   \hline 
		   \multicolumn{7}{l}{\textbf{Retrieval-Oriented Pretraining}}\\
		  % \midrule
		  \hline
		   SimCSE~\shortcite{gao-etal-2021-simcse} & 0.267	& 0.550 &		0.683	&0.290	& 0.379	& 0.434 \\
		   ICT~\shortcite{ict}  & 0.270 &	0.585&		0.697&	0.313&	0.374&	0.448\\
		   MLM~\shortcite{liu2019roberta}  & 0.302 &	0.600&	0.695& 0.304&	0.388 & 	0.458\\
		   TSDAE~\shortcite{tsdae} & 0.293 &	0.628	&	0.761 &	0.318 &	0.394 &	0.479 \\
		   Condenser~\shortcite{gao-callan-2021-condenser}  & 0.270 &	0.627 &		0.654	& 0.306	& 0.345	& 0.440 \\
		   Condenser (ours)  & 0.250 &	0.617 &	0.732	& 0.334	& 0.411	& 0.469 \\
		  % coCondenser$^\dagger$~\shortcite{cocondenser}  & 00.302	& 0.688	&	0.783 &	0.365 &	0.443 &	0.517\\
		  % \hline 
		   \hline 
		   \multicolumn{7}{l}{\textbf{In-Domain Generated Pseudo Labels}}\\
		   \hline 
		   QGen~\shortcite{genq} & 0.287 &	0.638	&	0.724	&0.330&	0.381&	0.472  \\
		   \multicolumn{5}{l}{GPL~\shortcite{wang2021gpl}} \\
		    w/ DistillBERT~\shortcite{sanh2019distilbert} & 0.328 &	0.664&		0.726&	0.345&	0.414&	0.495  \\
		    w/ TSDAE~\shortcite{tsdae} & 0.344 &	0.689	&	0.746 &	0.351 &	0.430	& 0.512  \\
		   \hline 
		   \multicolumn{7}{l}{\textbf{Reranking with Cross-Encoders,} considered as ``upper bound''~\shortcite{wang2021gpl}} \\
		   \hline
		   \multicolumn{5}{l}{Cross Encoder (MiniLM~\shortcite{wang2020minilm})} \\
		     w/ BM25 & 0.331 &	0.676	&	0.712&	0.368&	0.467&	0.511 \\
		    w/ TSDAE+GPL~\shortcite{wang2021gpl} & \bf 0.364 &	0.683	&	0.714	&  0.381&	\bf 0.483	& 0.525\\
		   \hline 
		   \multicolumn{5}{l}{\textbf{Our Method}} \\
		   \hline 
		   COCO-DR{\tiny{Base}} w/o iDRO & 0.302	& 0.688	&	0.785 &	0.365 &	0.443 &	 0.517 \\
		   COCO-DR\tiny{Base} & 0.307	&  0.709 & \bf 0.807&	0.370&	0.443 &	 0.527$^{\dagger}$ \\
		   COCO-DR\tiny{Large} & 0.329 & \bf  0.722 &  \bf 0.807 & \bf  0.393 & 0.482&	\bf 0.547$^{\dagger}$ \\
		   \bottomrule
		\end{tabular}
	%\end{center}
	}
	\caption{Comparison to domain adaptation methods on the BEIR tasks used in  \cite{wang2021gpl}. 
	$^{\dagger}$ indicates statistically significant results over the strongest baseline without using reranking models (GPL w/ TSDAE). 
	}
	\label{tab:uda_beir}
%\lgspace
\end{table}
\label{sec:coco_exp}
\begin{table*}[t]
\centering
% \small
\resizebox{\textwidth}{!}{%
\begin{tabular}{p{13.2cm}|l}
\toprule
\bf Target TREC-COVID Query   &\bf   {\marco} Nearest Query   \\ \hline 
does SARS-CoV-2 have any subtypes, and if so what are they? (\blue{+0.174}) & different types of hiv virus (\blue{+0.041}) \\ \hline 
how long can the coronavirus live outside the body (\blue{+0.057}) & how long does hep c live outside body (\blue{+0.056}) \\ \hline 
what are best practices in hospitals and at home in maintaining quarantine? (\blue{+0.045}) & define medical quarantine (\blue{+0.055}) \\ \hline 
% how long can the coronavirus live outside the body (\blue{+0.057}) & how long does hep c live outside body (\blue{+0.056}) \\ \hline 
is remdesivir an effective treatment for COVID-19 (\blue{+0.025}) & how are antiviral drugs effective in treating infection? (\blue{+0.031}) \\ \hline 
what are the impacts of COVID-19 among African-Americans that differ from the rest of the U.S. population? (\blue{+0.030}) & what ethnic group does sickle cell anemia affect (\blue{+0.026}) \\ 
\bottomrule
\end{tabular}%
}
\caption{Case study: Examples for nearest source queries of a target query in TREC-COVID and their performance gains after \model{} training. The \blue{number} in brackets denotes the nDCG@10 gain from iDRO.}
\label{tab:case}
\end{table*}

To further understand the benefit of continuous contrastive pretraining, we perform three experiments on it, including: (1) comparison with other unsupervised domain adaptation (UDA) approaches, (2) the correlations between pretraining and zero-shot, and (3) the pretrained sequence representations.

 \textbf{Comparison with UDA methods.} 
In Table~\ref{tab:uda_beir} we compare \model{} with methods besides dense retrieval on the five domain specific tasks used in the experimental settings of \citet{wang2021gpl}.\footnote{We omit BioASQ here as \citet{wang2021gpl} evaluated on its subset that is not public.}

\model{} outperforms all previous approaches, even those used a \textit{reranking} model upon first stage retrieval. The latter previously was viewed as the ``generalization upper bound'' since they use strong cross-encoder models that have access to term-level matching signals~\citep{wang2021gpl}.
Previous methods that conducted contrastive pretraining such as  ICT~\citep{ict} and SimCSE~\cite{gao-etal-2021-simcse} underperformed  simple BM25 in zero-shot retrieval.
These results corroborate the necessity of continuous contrastive learning.
% for the generalization ability.
% Next experiments further study why.

% \input{Figures/fig_coco}

 \textbf{Pretraining versus Zero-Shot.}  
In Figure~\ref{fig:coco_gain} we plot the reduction of the sequence contrastive learning loss after using COCO pretraining on BEIR corpora (versus pretraining only on MARCO corpus), as well as the corresponding zero-shot improvements on each BEIR task. There is a notable correlation between them. On BioASQ, COCO reduces contrastive loss by 50\% which yields 22\% gains in zero-shot.
Note that the pretrained models are fine-tuned \textit{solely} on MS MARCO, but they provide attributable gains in zero-shot afterward.

 \textbf{Pretrained Representations.}
Following \citet{wang2020hypersphere}, we use \emph{alignment} and \emph{uniformity} to illustrate the quality of learned representations on BEIR corpora (details in Appendix~\ref{app:alignment_uniformity}).
Figure~\ref{fig:uni_align} plots the results of \model{} on BEIR corpora with different pretraining components, before finetuning.
Without contrastive learning, {\condenser} representations are not well aligned, which results in degeneration on target tasks. Contrastive learning on MS MARCO does not capture the sequence representations on BEIR,  \model{} w/o COCO has low uniformity.
\model{} provides a balanced alignment and uniformity which leads to better generalization~\citep{wang2020hypersphere}.

\subsection{Influence of Implicit DRO}
\label{sec:idro_exp}
% The last experiment studies the benefits of iDRO. 

The assumption of iDRO is that it improves the model robustness on rare query clusters in \textit{source}, which helps generalize to unseen \textit{target}.
To verify this, we find MARCO query clusters closest to queries in each BEIR task (based on average dot product in \model{} embeddings). Then we plot the improvements of iDRO on BEIR tasks (zero-shot NDCG@10) and on their closest source clusters (training loss) in Figure~\ref{fig:idro_gain}. 

From the figure, we observe the connections between the two sides: iDRO improved the training loss on the majority (12 out of 18) of source query clusters closest to BEIR. 
Moreover, such improvements have been successfully propagated to the BEIR tasks, as there exists a clear positive correlations among the performance gain on the \marco{} and the corresponding target tasks. 
In Table~\ref{tab:case}, we show example query pairs with this connection on TREC-COVID to further support this argument. 
There are resemblance of the search intents between the source and target queries.
The improvements of iDRO on the source queries thus also lead to the gains on unseen queries in BEIR.

% To illustrate the effect of iDRO, we aim to establish the connection between the performance gain on different groups of {\marco} and BEIR tasks. 
% Specifically, for each BEIR task, we find it nearest group based on the average dot-product similarities on queries. 
% Then, Fig.~\ref{fig:idro} illustrates the correlation between the gain on each target task in BEIR and the gain on its nearest groups in {\marco} after adding iDRO.
% The result shows that compared with ERM, iDRO brings additional performance gains on groups that are more closed to the target tasks, and such gains can be further propagated to the target tasks to improve the zero-shot performance. 

% domain and {\marco} are close, indicating {\model} can well reduce the representation gap between the source and target domain. 
% Moreover, the performance gain in the target domain is also consistent with the gain on its nearest neighbor in {\marco}, which further consolidates the efficacy of iDRO for better robustness on rare query intents.
% \input{Figures/fig_idro}

\section{Conclusion}
\label{sec:conclusion}

\model{} improves ZeroDR accuracy by combating the distribution shifts using continuous contrastive learning and implicit distributionally robust optimization.
COCO helps models better capture the sequence representations of target corpora in pretraining.
Implicit DRO improves model robustness by reweighting query clusters in fine-tuning.

\model{} achieves strong zero-shot performance while maintaining a lightweight system with one unified model for all 18 target tasks.
Different than prior works that scaling up the DR model to billions of parameters (\emph{e.g.} CPT-text), we provide a more efficient and sustainable way to improve the zero-shot generalization ability. 
Our analyses observed clear correlations on \model{}'s ability to mitigate the distribution shifts and to generalize.
Better ZeroDR accuracy is observed on tasks where continuous contrastive learning has a lower pretraining loss, and where iDRO identifies  and improves source query clusters similar to target queries. 
% We hope our findings on the connection between distribution shifts and ZeroDR accuracy will inspire more studies in future research.

\section*{Limitations}
In this work, we propose {\model} to combat the distribution shift issue for zero-shot dense retrieval.  
Despite the strong performance of our two key designs (COCO and iDRO), we mainly verify their efficacy from their empirical performance on BEIR tasks. More theoretical analyses are required to gain deeper understandings of these two designs.
For COCO, more powerful tools are needed to establish the connection between contrastive pretraining and the performance on ZeroDR target tasks.
For iDRO, the key assumption is that the robustness over rare query clusters will lead to better zero-shot performance on target out-of-domain tasks. However, there are no theoretical groundings to connect these two terms for DR models. These analyses will go beyond our empirical observations and reveal the true inner workings of {\model}.

% Besides, our experiment results are all based on small or medium sized language models with million-scale parameters. Although \model{} performs on par with or better than previous ZeroDR systems with billions of parameters, it remains unknown to us how the benefit of \model{} scales with more parameters. Unfortunately, answering this question requires extensive computing resources which exceed our computational budget. It would be interesting to see how the performance of \model{} changes with more parameters.

\section*{Acknowledgements}
We would like to thank Ji Xin and Nandan Thakur for their help on getting access to non-public datasets of the BEIR benchmark.   
We also thank anonymous reviewers for their feedback. 
Yue Yu and Chao Zhang were partly supported by NSF IIS-2008334, IIS-2106961, and CAREER IIS-2144338.

% \section*{References}

% \newpage
% \balance

% {\small
% \bibliography{citation}
% \bibliographystyle{}
% }
\bibliography{anthology,citation,citation_modir}\bibliographystyle{acl_natbib}

% \input{Sections/checklist}

% \appendix

%%%%%%%%%%%%%%%%%%%%%%%%%%%%%%%%%%%%%%%%%%%%%%%%%%%%%%%%%%%%

\newpage
% \clearpage
\appendix
\section{Datasets Details}
\label{appx:datasets}
\begin{table*}[t!]
    \small
    \resizebox{\textwidth}{!}{\begin{tabular}{ l | l | l | c | c | c | c | c c c | c c }
        \toprule
         \multicolumn{1}{l}{\textbf{Split} ($\rightarrow$)} &
         \multicolumn{4}{c}{} &
         \multicolumn{1}{c}{\textbf{Train}}    &
         \multicolumn{1}{c}{\textbf{Dev}}    &
         \multicolumn{3}{c}{\textbf{Test}}   &
         \multicolumn{2}{c}{\textbf{Avg.~Word Lengths}} \\
         \cmidrule(lr){6-6}
         \cmidrule(lr){7-7}
         \cmidrule(lr){8-10}
         \cmidrule(lr){11-12}
           \textbf{Task ($\downarrow$)} &\textbf{Domain ($\downarrow$)} & \textbf{Dataset ($\downarrow$)} & \textbf{Title} & \textbf{Relevancy} & \textbf{\#Pairs} & \textbf{\#Query} & \textbf{\#Query} & \textbf{\#Corpus} & \textbf{Avg. D~/~Q } & \textbf{Query} & \textbf{Document} \\
         \midrule
    Passage-Retrieval    & Misc. & MS MARCO  & \xmark & Binary  & 532,761 &   ----  &   6,980   &   8,841,823      & 1.1 & 5.96  & 55.98  \\ \midrule[0.05pt] \midrule[0.05pt]
    Bio-Medical          & Bio-Medical & TREC-COVID   & \cmark & 3-level &   ----    &   ----  & 50     & 171,332   & 493.5& 10.60 & 160.77 \\
    Information          & Bio-Medical & NFCorpus    & \cmark & 3-level & 110,575 &  324  & 323    & 3,633     & 38.2 & 3.30  & 232.26 \\
    Retrieval (IR)       & Bio-Medical & BioASQ     & \cmark & Binary  & 32,916 & ---- & 500    & 14,914,602& 4.7  & 8.05  & 202.61 \\ \midrule
    Question             & Wikipedia  & NQ           & \cmark & Binary  & 132,803  &   ----  & 3,452 & 2,681,468 & 1.2  & 9.16  & 78.88  \\
    Answering       & Wikipedia  & HotpotQA     & \cmark & Binary  & 170,000 & 5,447 & 7,405  & 5,233,329 & 2.0  & 17.61 & 46.30  \\
     (QA)           &Finance& FiQA-2018   & \xmark & Binary  & 14,166  &  500  & 648    & 57,638    & 2.6  & 10.77 & 132.32 \\ \midrule
    Tweet-Retrieval      &Twitter& Signal-1M (RT)    & \xmark & 3-level &   ----    &   ----  & 97     & 2,866,316 & 19.6 & 9.30  & 13.93  \\ \midrule
    News      &News& TREC-NEWS      & \cmark & 5-level &   ----    &   ----  & 57     & 594,977 & 19.6 & 11.14  & 634.79  \\
    Retrieval      &News& Robust04  & \xmark & 3-level &   ----    &   ----  & 249   & 528,155 & 69.9 & 15.27  & 466.40  \\ \midrule
    Argument       & Misc. & ArguAna      & \cmark & Binary  &   ----    &   ----  & 1,406  & 8,674     & 1.0  & 192.98& 166.80 \\
    Retrieval   & Misc. & Touch\'e-2020 & \cmark & 3-level &   ----    &   ----  & 49     & 382,545   & 19.0 & 6.55  & 292.37 \\ \midrule
    Duplicate-Question   &StackEx.& CQADupStack   & \cmark & Binary  &   ----    &   ----  & 13,145 & 457,199   & 1.4  & 8.59  & 129.09 \\
    Retrieval            & Quora &  Quora        & \xmark & Binary  &   ----    & 5,000 & 10,000 & 522,931   & 1.6  & 9.53  & 11.44  \\ \midrule
    Entity-Retrieval     & Wikipedia  &  DBPedia      & \cmark & 3-level &   ----    &   67  & 400    & 4,635,922 & 38.2 & 5.39  & 49.68  \\ \midrule
    Citation-Prediction  & Scientific&  SCIDOCS       & \cmark & Binary  &   ----    &   ----  & 1,000  & 25,657    & 4.9  & 9.38  & 176.19 \\ \midrule
                         & Wikipedia  &  FEVER       & \cmark & Binary  & 140,085 & 6,666 & 6,666  & 5,416,568 & 1.2  & 8.13  & 84.76  \\ 
    Fact Checking        & Wikipedia  & Climate-FEVER  & \cmark & Binary  &   ----    &   ----  & 1,535  & 5,416,593 & 3.0  & 20.13 & 84.76  \\
                         & Scientific & SciFact     & \cmark & Binary  &   920      &   ----  &  300   & 5,183     & 1.1  & 12.37 & 213.63  \\
    \bottomrule
    \end{tabular}}
    \caption{Statistics of datasets in the BEIR benchmark. The table is taken from the original BEIR benchmark paper~\cite{beir}. \vspace{-3mm}}
    \label{tab:dataset_stats}
\end{table*}

Target domain datasets used in our experiments are collected in the BEIR benchmark~\citep{beir}\footnote{\url{https://github.com/beir-cellar/beir}} and include the following domains:
\begin{itemize}[leftmargin=*]
    \item Bio-Medical Information Retrieval: TREC-COVID~\citep{treccovid}, NFCorpus~\citep{nfcorpus}, and BioASQ~\citep{bioasq}.
    \item Open-domain Question Answering (QA): HotpotQA~\citep{hotpotqa},  NQ~\citep{nq}, and FiQA~\citep{fiqa}.
    \item Argument Retrieval: Webis-Touch\'e2020~\citep{touche} and ArguAna~\citep{arguana}.
    \item News Retrieval: TREC-NEWS~\citep{trecnews} and Robust04~\citep{robust04}.
    \item Tweet Retrieval: Signal-1m~\citep{signal1m}.
    \item Duplicate Question Retrieval: Quora~\citep{beir} and CQADupStack~\citep{cqadupstack}.
    \item Entity Retrieval: DBPedia~\citep{dbpedia}
    \item Citation Prediction:  SCIDOCS~\citep{scidocs}
    \item Fact Checking: SciFact~\citep{scifact}, FEVER~\citep{fever}, and Climate-FEVER~\citep{climatefever}
\end{itemize}
We list the statistics of the BEIR benchmark in Table~\ref{tab:dataset_stats}.

%  The 18 English  zero-shot evaluation datasets come from 9 heterogeneous retrieval tasks, including bio-medical information retrieval, question answering, tweet retrieval, news retrieval, argument retrieval, duplicate question retrieval, citation prediction, and fact checking. 

\paragraph{Metric} To measure the effectiveness of search algorithms or retrieval models, the benchmark uses Normalized Discounted Cumulative Gain (nDCG@10) \cite{ndcg} as the evaluation metric.  The higher value indicates better performance.
% We will give the definition of the metric in the following.

% Given top $k$ retrieved documents $\{d_1,d_2,\ldots,d_k\}$ with their relevance score $\{r_1,r_2,\ldots,r_k\}$ for a query, the traditional formula of discounted cumulative gain (DCG) accumulated at a particular rank position $k$ is defined in Equation \ref{eq:dcg}, where $r_i$ is 1 if $d_i$ is the ground truth otherwise 0.
% \begin{equation}
% \label{eq:dcg}
%     DCG@K = \sum_{i=1}^K\frac{r_i}{log_2(i+1)}
% \end{equation}

% Since the length of ground truth list depends on the query, using DCG to compare the performance of retrieval models from one query to the next cannot be consistently achieved. Therefore, the discounted cumulative gain is normalized  (nDCG) as:
% \begin{equation}
% \label{eq:ndcg}
%     {nDCG}@K = \frac{DCG@K}{IDCG@K}
% \end{equation}
% where IDCG@K is the DCG@K score for the list of relevant documents (ordered by their relevance) in the corpus up to position $k$. Since IDCG@K producs the maximum possible DCG through position $k$, the value of nDCG@K is in the range 0 to 1.

\section{Baselines}
\label{appx:baseline}
We use the baselines from the current BEIR leaderboard~\citep{beir} and recent papers. For the main experiments, the baselines can be divided into four groups: dense retrieval,  dense retrieval with generated queries\footnote{We separate them from dense retrieval since they usually rely on Seq2seq models to generate pseudo query-document pairs, and they train a model for each dataset \emph{independently} instead of using a single model for all datasets.}, lexical retrieval, and late interaction. 

\subsection{Baselines for Main Experiments}

\paragraph{Dense Retrieval}
For dense retrieval, the baselines are the same dual-tower model as ours. We consider \textbf{DPR}~\cite{dpr}, \textbf{ANCE}~\cite{ance}, 
\textbf{Contriever}~\cite{contriever}, 
and two recently-proposed giant model, namely  \textbf{GTR}~\cite{gtr} and \textbf{CPT-text}~\cite{cpt} in this paper.
\begin{itemize}[leftmargin=*]
    \item \textbf{DPR} uses a single BM25 retrieval example and in-batch examples as hard negative examples to train the model. Different from the original paper~\cite{beir} that train the DPR on QA datasets, we train DPR on MS MARCO~\cite{msmarco} Dataset for \emph{fair comparison}. Notice that this also lead to better results  according to \citet{modir}. 
    \item \textbf{ANCE}  constructs hard negative examples from an ANN index of the corpus. The hard negative training instances are updated in parallel during fine-tuning of the model. The model is a RoBERTa~\cite{liu2019roberta} model trained on MS MARCO for 600k steps.
    % \item \textbf{TAS-B} is trained with Balanced Topic Aware Sampling using dual supervision from a cross-encoder and a ColBERT model~\citep{colbert}. The model is trained with a combination of a pairwise Margin-MSE~\cite{tasb} loss and an in-batch negative loss function.
    
    \item \textbf{Contriever}  conducts unsupervised contrastive pretraining with data augmentations and momentum queues on Wikipedia and CC-Net~\cite{ccnet} corpora for 500k steps.

     \item \textbf{GTR} initializes the dual encoders from the T5 models~\cite{raffel2019t5}. It is first pre-trained on
     Community QA\footnote{Unfortunately, this corpus is not publicly available.} with 2 billion question-answer pairs then fine-tuned on NQ and MS Marco dataset.

     \item \textbf{CPT-text} initializes with the large GPT models~\cite{gpt3}, and pre-trained on web-scale Internet data with neighboring pieces of text as positive pairs for the contrastive objective. 
\end{itemize}

% \cite{}
\paragraph{Dense Retrieval with Generated Queries}
\begin{itemize}[leftmargin=*]
    \item \textbf{GenQ} first fine-tunes a T5-base~\cite{raffel2019t5} model on MS MARCO for 2 epochs and then generate 5 queries for each passage as additional training data for the target domain to continue to fine-tune the TAS-B~\cite{tasb} model.
    \item \textbf{GPL} is a recent work that improve the perforance of GenQ with cross-encoder reranking.  It first generates queries for documents from the target domain, then use an additional cross-encoder~\cite{wang2020minilm} to rank each (query, document)-pair and then train a dense retrieval model on these generated, pseudo-labeled queries\footnote{In the original paper, they have tried on multiple backbones including DistillBERT~\cite{sanh2019distilbert}, TSDAE~\cite{tsdae}, TAS-B~\cite{tasb} for evaluations, and we select the best model that based on TAS-B for comparison in our main experiments.}. 
\end{itemize}
    
%  \textbf{GenQ}~\cite{genq}
\paragraph{Lexical Retrieval}
Lexical retrieval is a score function for token matching calculated between two high-dimensional sparse vectors with token weights. 
\begin{itemize}[leftmargin=*]
\item \textbf{BM25}~\cite{bm25} is the most commonly used lexical retrieval function. We use the BM25 results reported in \citet{beir} for comparison.
% \item \textbf{DocT5query}~\cite{doct5query} is a popular document expansion technique using a T5 (base) model trained on MS MARCO to generate synthetic queries and append them to the original document for lexical search. 
\end{itemize}

\paragraph{Late Interaction}
We also consider a late interaction baseline, namely \textbf{ColBERT}~\cite{colbert}. The model computes multiple contextualized embeddings for each token of queries and documents, and then uses a maximum similarity function to retrieve relevant documents. This type of matching requires significantly more disk space for indexes and has a higher latency.

% \paragraph{Re-ranking}
% Re-ranking based approaches use the output of a first-stage retrieval system (\eg BM25), and then re-rank the retrieved documents using a cross-encoder~\cite{nogueira2020passage}. In this paper, we use the \textbf{BM25+CE} baseline implemented in \citet{beir} that uses BM25 to retrieve top-100 documents and a 6-layer MiniLM~\cite{minilm} model to further re-rank the recalled documents. 
\begin{table*}[t]
%   \small
% \vspace{-1.5ex}
  \centering
  \resizebox{0.8\linewidth}{!}{%
    \begin{tabular}{l|c|c|c|c}
\toprule
\bf \multirow{2}{*}{Dataset ($\downarrow$)} & {\bf Query Intent}      & {\bf Document Lexical} & {\bf ANCE (BERT$_\text{Base}$)} & {\bf ANCE ({\cocondenser})} \\
& {\bf Similarity} & {\bf Similarity} & {\bf v.s. BM25} & {\bf v.s. BM25} \\ \hline
TREC-COVID     &
0.4845 &	0.2789 &-0.002 &	+0.102
 \\
BioASQ      & 
0.4380 &	0.2806&-0.159&	-0.124 \\
NFCorpus             & 
0.2367&	0.2426  &-0.088&	+0.001\\
NQ          & 
0.5127&	0.5092 &+0.117	&+0.174 \\
HotpotQA         &
0.5078&	0.3275&-0.147&	-0.019	\\
FiQA-2018    & 
0.4950 &	0.3721 &+0.059	&+0.067	 \\
Signal-1M          & 
0.1708&	0.3334&-0.081&	-0.056 \\
TREC-NEWS                &
0.2280&	0.4194 &-0.016&	+0.002\\
Robust04      &
 0.6656&	0.4323 &-0.016&	+0.008\\
ArguAna          & 
0.1690&	0.3421 &+0.001	&+0.046 \\
Touché-2020      & 
0.0391&	0.3785 &-0.127&	-0.127 \\
Quora                & 
0.5629&	0.4141 &+0.063	&+0.071 \\
DBPedia-entity              & 
0.2235&	0.3189 &-0.032&	+0.051 \\
SCIDOCS            & 
0.1636&	0.2945 &-0.036&	-0.008 \\
Fever            & 
0.1621&	0.3689 &-0.084&	-0.002  \\
Climate-Fever         & 
0.1732&	0.3689 &-0.015&	-0.014\\
SciFact               & 
0.1809&	0.2335 &-0.158&	-0.092\\
CQADupStack         & 
0.4254&	0.3196 &-0.003&	+0.043 \\ 
% Avg w/o MS Marco                                      & \multicolumn{1}{c|}{0.423}          & {\underline{0.434}}    & \multicolumn{1}{c|}{0.237} & \multicolumn{1}{c|}{0.389} & \multicolumn{1}{c|}{{\underline{0.415}}} & \multicolumn{1}{c|}{0.410} & 0.431          & \multicolumn{1}{c|}{0.416}    & \multicolumn{1}{c|}{0.445}     & \multicolumn{1}{c|}{0.453}          & \textbf{0.458} \\ 
\bottomrule
\end{tabular}
  }
%   \vspace{-1ex}
  \caption{Detailed statistics for (1) query intent similarity and document lexical similarity between {\marco} and BEIR tasks (2) the performance gap between ANCE starting from BERT$_\text{base}$ and coCondenser and BM25. The positive value indicates ANCE performs better than BM25. 
  }
  \label{tab:sim_qd_statistics}
\end{table*}

\subsection{Additional Domain Adaptation Baselines}
We further compare {\model} with additional baselines focus on domain adaptation to specialized domains including \textbf{UDALM}~\cite{udalm}, \textbf{MoDIR}~\cite{modir}, \textbf{SimCSE}~\cite{gao-etal-2021-simcse}, \textbf{ICT}~\cite{ict}, \textbf{MLM}~\cite{liu2019roberta}, \textbf{TSDAE}~\cite{tsdae}, and \textbf{Condenser}~\cite{gao-callan-2021-condenser}.   
% \textbf{coCondenser}~\cite{cocondenser}  . 
Note that these models are first pre-trained on the target corpus and then fine-tuned on the MS MARCO dataset.
\begin{itemize}[leftmargin=*]
\item \textbf{UDALM} is a domain adaptation method that originally designed for sentiment analysis. It applies the multi-task training to jointly learn from the target task and the MLM task.
\item \textbf{MoDIR} is a momentum-based method to ensure stable and
efficient adversarial learning for domain adaptation.
\item \textbf{SimCSE} is a simple approach proposed for sentence similarity calculation. Specifically, it regards the document text twice with different dropout as the positive sample pairs to enable contrastive learning.
\item \textbf{ICT} selects one sentence from a whole document as the pseudo query to that document for pre-training.
\item \textbf{MLM} random masks 15\% tokens in a text and designs a cloze-style test for pre-training the model.
\item \textbf{TSDAE} leverages an additional denoising autoencoder to pre-train the dense retriever model with 60\% random tokens deleted in the input document.
\item \textbf{Condenser} improves the representation of \texttt{[CLS]} token by enforcing it to aggregate with the token embedding. In this way, the head model can then condition on late \texttt{[CLS]} to make LM predictions to enforce \texttt{[CLS]} to capture the global meaning of the input text.
% \item \textbf{coCondenser} complements the Condenser pre-training  via adding an additional contrastive learning objective. Specifically, it regards two spans from the same document as the positive pairs.
\end{itemize}

\section{Details for Similarity Calculation}
In this section, we provide more details on how to calculate the distribution shifts between the source training task ({\marco}) and the zero-shot target tasks (BEIR). 
We first define the types of queries used in Section \ref{sec:motivation_shift}, and then give more details about the calculation of the weighted Jaccard similarity~\cite{jaccard} used in this study.

\subsection{Types of Queries}
\label{app:query_type}
We adopt the same  method as \cite{ren2022thorough} to partition the training queries into 9 types: for queries starting with the following 7 words, 'what', `when', `who', `how', `where', `why', `which', they fall into the corresponding category. 
Besides,  queries starting with the first word \texttt{is/was/are/were/do/does/did/have/has/had/ 
should/can/could/would/am/small}',  are classified as Y/N queries. 
The rest of the queries belong to declarative queries.

\subsection{Calculation of Weighted Jaccard Similarity}
We follow \cite{beir} to use the weighted Jaccard similarity $J(S, T)$ to measure the unique word overlap for all words present in the source dataset $S$ and the target dataset $T$. 

Denote $S_k$ as the frequency of word $k$ in the source dataset $S$ and $T_k$ for the target dataset $T$ respectively. The weighted Jaccard similarity $J(S, T)$ between $S$ and $T$ is defined as:
\begin{equation}
J(S, T)=\frac{\sum_k \min \left(S_k, T_k\right)}{\sum_k \max \left(S_k, T_k\right)},
\end{equation}
where the sum is over all unique words $k$ present in dataset $S$ and $T$.

\section{Statistics for Query and Document Similarities}
Table \ref{tab:sim_qd_statistics} lists the exact pairwise weighted Jaccard similarity between {\marco} and different BEIR tasks. 
For tasks comes from biomedical domains (e.g. BioASQ, NFCorpus) and scientific domains (e.g. SCIDOCS, SciFact), the lexical overlap between them and {\marco} is small. For these datasets, ANCE can hardly outperform BM25. 
On the other hand, for those tasks which ANCE outperforms BM25 by a wide margin (e.g. NQ, Quora), they tend to have a larger weighted Jaccard similarity score with MS MARCO.

\section{Details of iDRO}
\label{app:idro}
This section exhibits the details for deriving the optimal weight $\boldsymbol{\omega}^{(t)}$ for the training step $t$. Note that the overall objective can be expressed as 
% \begin{equation}
\begin{align}
&\min _{\omega^{(t)}} \ \ell_{\text{g}} + \tau \mathcal{D}_{\text{KL}}(\boldsymbol{\omega}^{(t)}||\boldsymbol{\omega}^{(t-1)}), \\
&\text{ s.t.} \quad \sum_{i=1}^{K}{\omega_i^{(t)}}=1,
\label{eq:app_constrain_reg}
\end{align}
% \end{equation}
where $\tau$ is the temperature to control the strength of the regularization. Then, the KKT conditions can be expressed as 
% \begin{equation}
\begin{align}
\cL &=-\sum_{i=1}^{K}\sum_{j=1}^{K}\omega_i\alpha_{i}\alpha_{j}\left(\nabla_{\theta}\ell_i(\theta)\right)^{\texttt{T}}\nabla_{\theta}\ell_j(\theta) \\
&+\tau \sum_{i=1}^{K} \left(\log \left(\frac{\omega^{(t)}}{\omega^{(t-1)}}\right)-1\right) \\
&+\gamma\left(\sum_{i=1}^{K} \omega^{(t)}_{i}-1\right)
\label{eq:app_lagrangian}
\end{align}
% \end{equation}

\noindent Setting the corresponding gradients to 0 gives the global optimum as

\begin{small}
\begin{equation}
\frac{\partial \cL}{\partial {\omega}_{i}^{(t)}}=-\sum_{j=1}^{K}r_{ij} +\tau \log \left(\frac{\omega^{(t)}}{\omega^{(t-1)}}\right)+\hat{\gamma}=0; 
\label{eq:kkt_equation}
\end{equation}
\end{small}
\begin{small}
\begin{equation}
\sum_{i=1}^{K}{\omega}_{i}^{(t)} = 1, 
\label{eq:sum_equation}
\end{equation}
\end{small}

\noindent where $$r_{ij}=\sum_{i=1}^{K}\alpha_{i}\alpha_{j}\left(\nabla_{\theta}\ell_i(\theta)\right)^{\texttt{T}}\nabla_{\theta}\ell_j(\theta),$$  $$\hat{\gamma}=\gamma+\tau.$$ 
From the above Eqn.~\ref{eq:kkt_equation}, we have

% \begin{small}
\begin{equation}
\omega^{(t)}_i = \omega^{(t-1)}_i \exp\left(\frac{1}{\tau}\left(\sum_{j=1}^{K}r_{ij}-\hat{\gamma}\right)\right).
\label{eq:kkt_omega}
\end{equation}
% \end{small}

\noindent By plugging the Eqn.~\ref{eq:kkt_omega} to Eqn.~\ref{eq:sum_equation}, we obtain
% \begin{small}
\begin{equation}
\exp \left(\frac{\hat{\gamma}}{\tau}\right) 
= \sum_{i=1}^{K} \exp\left(\frac{1}{\tau}\sum_{i=1}^{K}\omega^{(t-1)}_i r_{ij}\right).
\label{eq:final_gamma}
\end{equation}
% \end{small}

\noindent Finally, by combining the Eqn.~\ref{eq:kkt_omega} and Eqn.~\ref{eq:final_gamma}, the weight for $i$-th group can be expressed as 
% \begin{small}
\begin{equation}
\omega^{t*}_i = \frac{\omega^{(t-1)}_i \exp\left(\frac{1}{\tau}\sum_{j=1}^{K}r_{ij}\right)}{\sum_{i=1}^{K}\omega^{(t-1)}_i \exp\left(\frac{1}{\tau}\sum_{j=1}^{K}r_{ij}\right)}.
% &= \frac{\omega^{(t-1)}_i \exp\left(\frac{1}{\tau} \sum_{j=1}^{N}r_{ij}\right)}{\sum_{i=1}^{N}\omega^{(t-1)}_i \exp\left(\frac{1}{\tau} \sum_{j=1}^{N}r_{ij} \right)},
% \end{aligned}
\label{eq:omega}
\end{equation}
% \end{small}

\section{Comparision with GroupDRO}
\label{app:group_dro}
\begin{table}[t]
%   \small
% \vspace{-1.5ex}
  \centering
  \resizebox{0.8\linewidth}{!}{%
    \begin{tabular}{l|c|c}
\toprule
\bf Dataset ($\downarrow$)& \multicolumn{1}{c|}{\bf COCO-DR}      & \multicolumn{1}{c}{\textbf{GroupDRO}~\shortcite{Sagawa2020Distributionally}}  \\ \hline
TREC-COVID     &
0.789 &	\textbf{0.793} \\
BioASQ      & 
\textbf{0.429} & 0.411 \\
NFCorpus             & 
\textbf{0.355} & 0.352 \\
NQ          & 
\textbf{0.505} &	 0.494\\
HotpotQA         &
\textbf{0.616} & 0.609	\\
FiQA-2018    & 
\textbf{0.307} & 0.300	 \\
Signal-1M          & 
0.271 &	\textbf{0.274} \\
TREC-NEWS                &
0.403 &	\textbf{0.408} \\
Robust04      &
\textbf{0.443} &	0.438 \\
ArguAna          & 
0.493 & 0.493 \\
Touché-2020      & 
0.238 &	\textbf{0.243} \\
Quora                & 
\textbf{0.867} &	0.866 \\
DBPedia-entity              & 
\textbf{0.391} & 0.390 \\
SCIDOCS            & 
0.160 & \textbf{0.162} \\
Fever            & 
\textbf{0.751} & 	0.746 \\
Climate-Fever         & 
{0.211} & 0.211 \\
SciFact               & 
0.709 & \textbf{0.712} \\
CQADupStack         & 
\textbf{0.370} & 0.367 \\ \hline
Avg           & 
\textbf{0.462} & 0.459 \\
% Avg w/o MS Marco                                      & \multicolumn{1}{c|}{0.423}          & {\underline{0.434}}    & \multicolumn{1}{c|}{0.237} & \multicolumn{1}{c|}{0.389} & \multicolumn{1}{c|}{{\underline{0.415}}} & \multicolumn{1}{c|}{0.410} & 0.431          & \multicolumn{1}{c|}{0.416}    & \multicolumn{1}{c|}{0.445}     & \multicolumn{1}{c|}{0.453}          & \textbf{0.458} \\ 
\bottomrule
\end{tabular}
  }
  \vspace{-1ex}
  \caption{Comparision between iDRO and GroupDRO~\cite{Sagawa2020Distributionally}. {\model} achieves better performance on the majority of BEIR tasks.
  }
  \label{tab:group_dro}
\end{table}
We further compare iDRO with GroupDRO~\cite{Sagawa2020Distributionally}, which assigns higher weights to groups with higher training loss. 
Note that GroupDRO requires gold labels for group assignments which is unavailable for ZeroDR. 
To adopt GroupDRO in our settings, we use the cluster information derived from K-means clustering as group labels, which is the same as  \cite{sohoni2020no}. To ensure fair comparison, we use the model after COCO pretraining as initialization, and use GroupDRO to reweight different groups during fine-tuning the model on {\marco}.

Table~\ref{tab:group_dro} shows the performance of GroupDRO on BEIR tasks.
From the results, we find that although GroupDRO achieves better performance on some specific tasks (e.g. TREC-COVID and SciFact), it fails to perform well on the majority of tasks, especially for general-domain datasets such as NQ, HotpotQA and Fever. 
This is because during GroupDRO training, it assigns higher weights for large-loss groups while neglecting other groups. As a result, although it will lead to better worse-group performance, it cannot improve the average performance. 
In contrast, iDRO leverages gradient similarities to dynamically reweight different groups to avoid sacrificing the average performance on all tasks.

\begin{figure*}[t]
\centering
\vspace{-3mm}
\subfigure[TREC-COVID]{
    \includegraphics[width=0.331\textwidth]{Figures/abla_trec_covid.pdf}
            \label{fig:ablation_treccovid}
        }\hfill   \hspace{-6mm}
    %  \caption{}
\subfigure[BioASQ]{
    \includegraphics[width=0.331\textwidth]{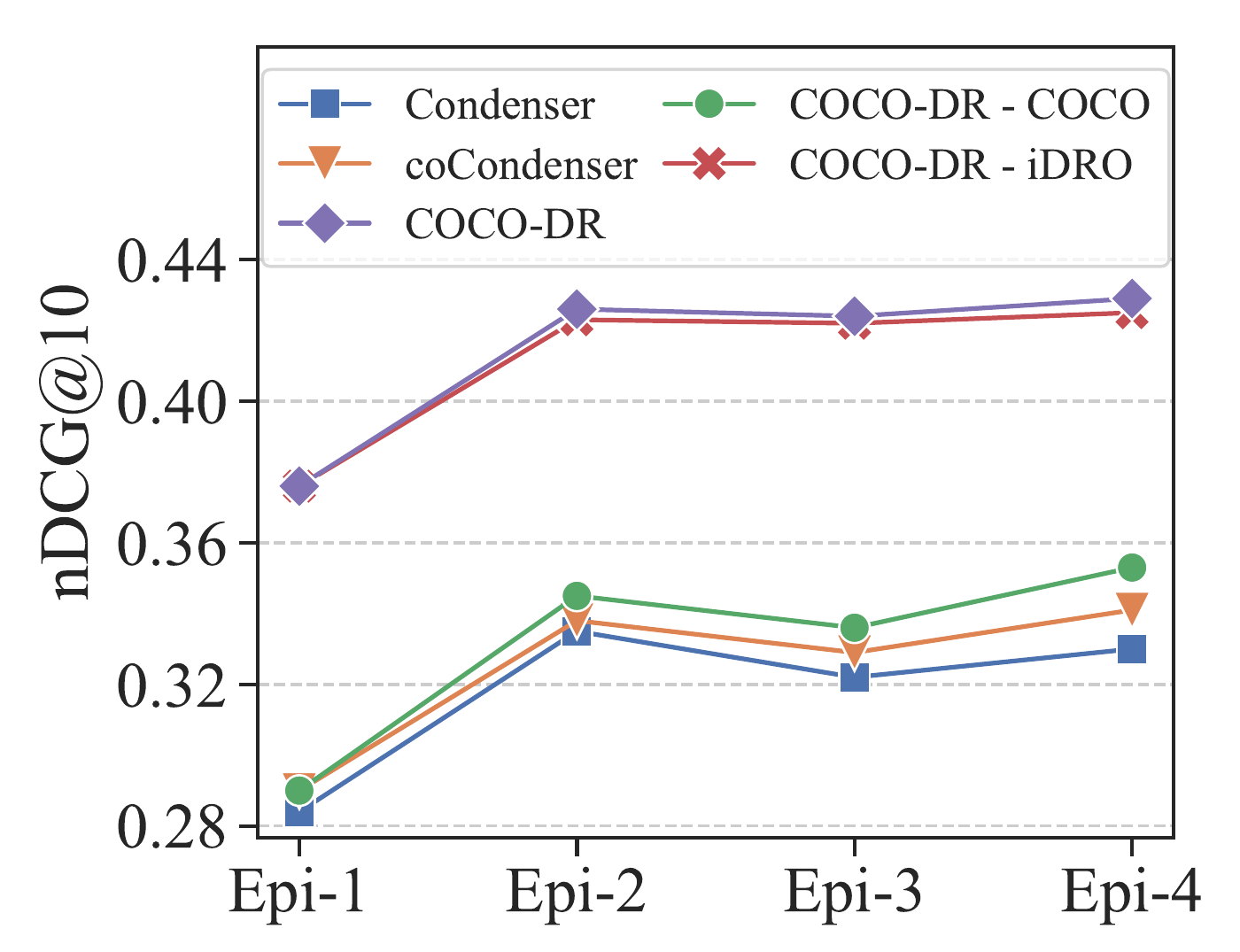}
            \label{fig:ablation_bioasq}
        }\hfill   \hspace{-6mm}
    %  \caption{BioASQ}
\subfigure[FiQA-2018]{
    \includegraphics[width=0.331\textwidth]{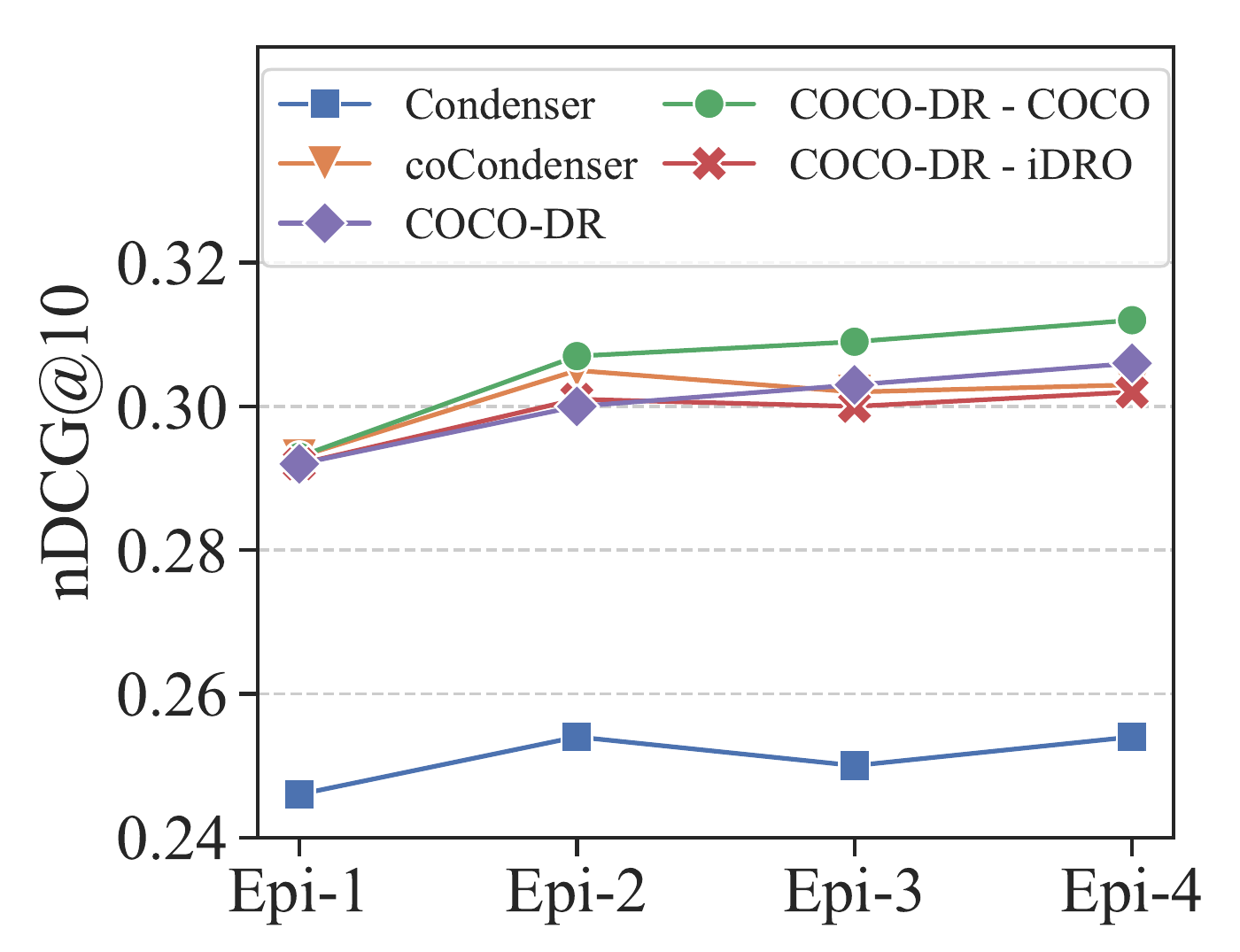}
            \label{fig:ablation_fiqa}
        }\hfill 
    %  \centering
    %  \hspace{-17pt}
    %  \includegraphics[width=\textwidth]{Figures/abla_fiqa2018.pdf}
    %  \caption{}
\vfill
\vspace{-3mm}
\subfigure[CQADupStack]{
    \includegraphics[width=0.331\textwidth]{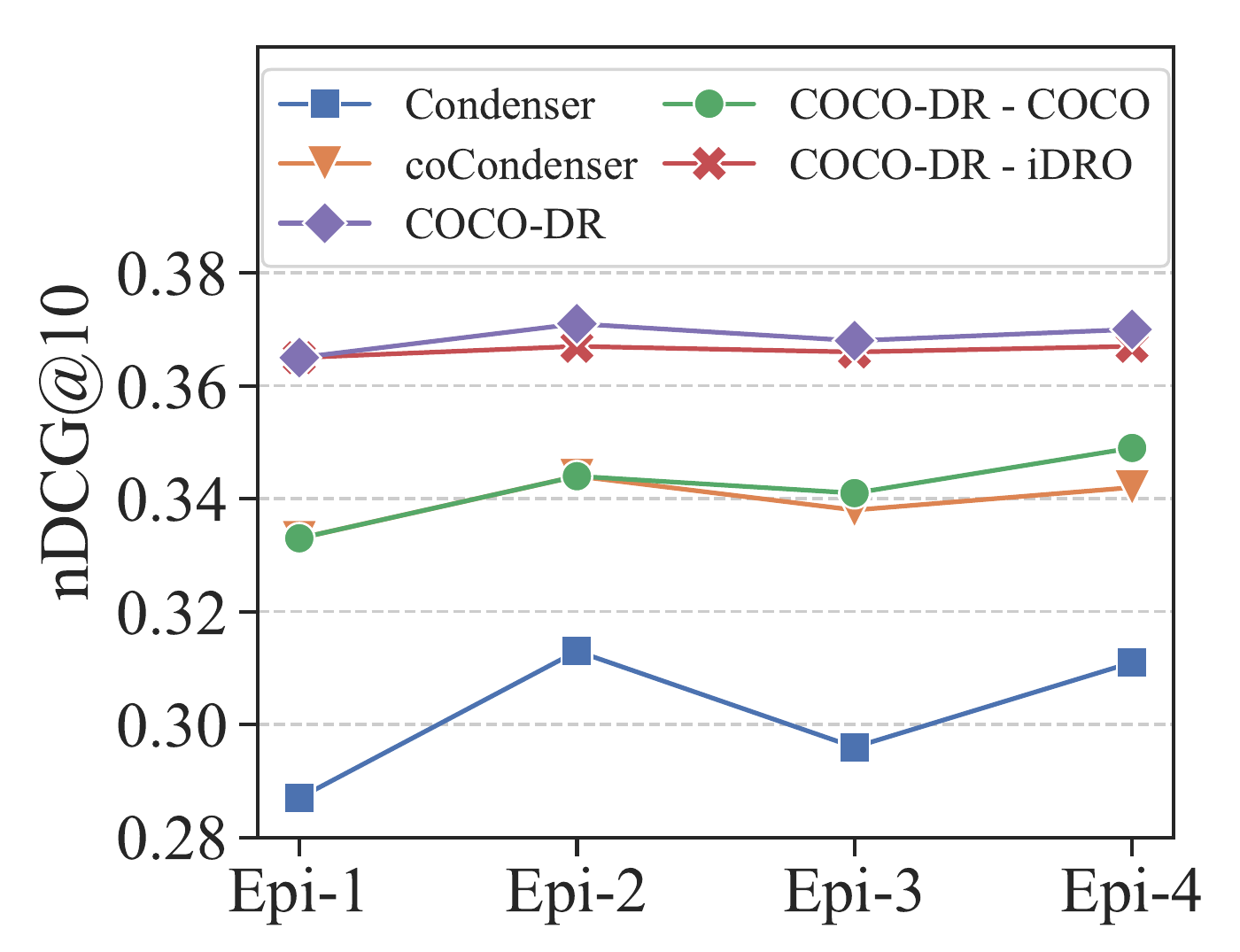}
            \label{fig:ablation_cqad}
        }\hfill  \hspace{-6mm}
    %  \caption{}
\subfigure[SciFact]{
    \includegraphics[width=0.331\textwidth]{Figures/abla_scifact.pdf}
            \label{fig:ablation_scifact}
        }\hfill  \hspace{-6mm}
    %  \caption{}
\subfigure[Robust04]{
    \includegraphics[width=0.331\textwidth]{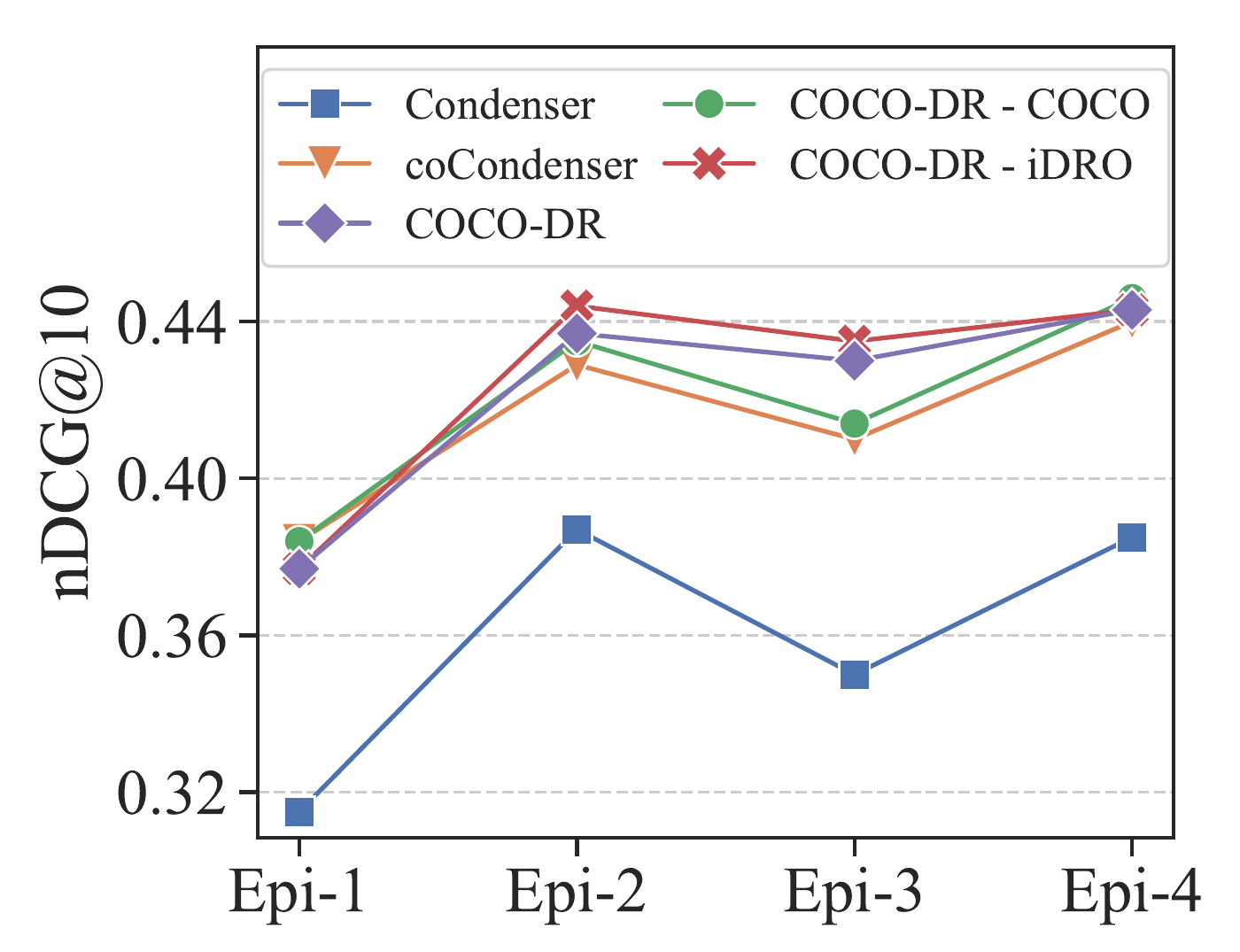}
            \label{fig:ablation_robust04}
        }\hfill  
% \begin{subfigure}[]{0.24\textwidth}
%      \centering
%      \includegraphics[width=\textwidth]{figs/tsne-repeat1-step30k.pdf}
%      \caption{w/o Mom. (30k)}
% \end{subfigure}
% \begin{subfigure}[]{0.24\textwidth}
%      \centering
%      \includegraphics[width=\textwidth]{figs/tsne-repeat1-step50k.pdf}
%      \caption{w/o Mom. (50k)}
% \end{subfigure}
% \vspace{-2ex}
\caption{The performance of {\model} and its variants over different training stages on 6 of BEIR tasks.
}
\label{fig:beir_dynamics_full}
\end{figure*}

\section{Performance on Different Training Stages of {\model}}
\label{app:dynamics}
Figure~\ref{fig:beir_dynamics_full} exhibits the performance on  different episodes on six BEIR tasks from different domains, used in \citep{wang2021gpl}. 
From the results, we observe that COCO is more beneficial for the biomedical domains than others such as news and finance. 
The more significant gain is mainly due to the limited overlap between biomedical corpus and {\marco}, as well as the extremely large size of the biomedical corpora. 
For other two tasks (Robust04 and FiQA-2018), the DR models can already achieve better or comparable performance compared with BM25 when finetuning on {\marco} only, which indicates the distribution shift issue is not severe on these datasets. 
Therefore, the relative gain of COCO on them is smaller. 

For the iDRO part, it provides additional performance gains on 5 of 6 datasets. 
As these datasets are all domain specific text retrieval tasks~\cite{wang2021gpl}, the results justify the benefits of iDRO for improving the DR model's performance on unseen target queries.

% \section{Full Evaluation Results on BEIR Benchmark}

\section{Calculation of Alignment and Uniformity}
\label{app:alignment_uniformity}

Recently, \citet{wang2020hypersphere} propose two terms, namely \emph{alignment} and \emph{uniformity} to measure the quality of
representations. 
In particular, we denote the whole data distribution as $p_{\text{data}}$ and the distribution of positive pairs as $p_{\text{pos}}$. 
Then, the two metrics can be calculated as 
\begin{align}
% \begin
\ell_{\text{align}} & \triangleq %\underset{}
{\mathbb{E}}_{(x, x^{+}) \sim p_{\text{pos}}}\|f(x)-f(x^{+})\|^{2}, \\
\ell_{\text{uniform}} & \triangleq \log %\underset{}
{\mathbb{E}_{(x, y) \stackrel{\text{i.i.d.}}{\sim} p_{\text{data}}}} e^{-2\|f(x)-f(y)\|^{2}}.
% \end{gathered}
\end{align}
Notably, \emph{alignment} is the expected distance between the representations of positive text pairs, and \emph{uniformity} measures how well the text  representations are uniformly distributed~\cite{gao-etal-2021-simcse}. 
In our experiments, we use the code released by the original authors to calculate these two metrics.\footnote{Link: \url{https://github.com/SsnL/align_uniform}}

\label{sec:appendix}

\end{document}